\documentclass[10pt,twocolumn,letterpaper]{article}

\usepackage[pagenumbers]{cvpr}

\makeatletter
\DeclareRobustCommand\onedot{\futurelet\@let@token\@onedot}
\def\@onedot{\ifx\@let@token.\else.\null\fi\xspace}

\def\eg{\emph{e.g}\onedot} 

\def\ie{\emph{i.e}\onedot}

\makeatother

\usepackage[accsupp]{axessibility}

\usepackage{amsmath}
\usepackage{amssymb}
\usepackage{amsfonts}

\usepackage{graphicx}
\usepackage{subcaption}
\usepackage{caption}
\usepackage{tikz}
\usetikzlibrary{shapes.geometric, arrows.meta, positioning, fit, backgrounds, calc}

\usepackage{booktabs}
\usepackage{multirow}
\usepackage{tabularray}
\UseTblrLibrary{booktabs}

\usepackage[dvipsnames]{xcolor}   

\definecolor{turquoise}{cmyk}{0.65,0,0.1,0.3}
\definecolor{purple}{rgb}{0.65,0,0.65}
\definecolor{dark_green}{rgb}{0,0.5,0}
\definecolor{orange}{RGB}{255,127,0}
\definecolor{red}{rgb}{0.8,0.2,0.2}
\definecolor{darkred}{rgb}{0.6,0.1,0.05}
\definecolor{blueish}{rgb}{0.0,0.3,0.6}
\definecolor{blue}{rgb}{0,0.3,1}
\definecolor{pink}{rgb}{1,0,1}
\definecolor{greyblue}{rgb}{0.25,0.25,1}
\definecolor{light_gray}{gray}{0.95}
\definecolor{light-green}{rgb}{0.82,0.94,0.75}
\definecolor{Alto}{rgb}{0.858,0.858,0.858}
\definecolor{green}{rgb}{0,0.8,0.2}

\usepackage{algorithm}
\usepackage{algpseudocode}

\usepackage{xspace}
\usepackage{pifont}          

\newcommand{\figref}[1]{Fig.~{\color{red}{\ref{#1}}}}
\newcommand{\tabref}[1]{Tab.~{\color{red}{\ref{#1}}}}

\renewcommand{\algref}[1]{Algorithm~{\color{red}{\ref{#1}}}}

\newcommand{\gray}[1]{{\color{gray}#1}}
\newcommand{\blue}[1]{{\color{blue}#1}}
\newcommand{\green}[1]{{\color{green}#1}}
\newcommand{\red}[1]{{\color{red}#1}}

\newcommand{\paragrapht}[1]{\noindent\textbf{#1}}
\newcommand{\cmark}{\ding{51}}
\newcommand{\xmark}{\gray{\ding{55}}}

\newcommand{\method}{Think-Compress-Answer}
\newcommand{\methodshort}{TCA}

\usepackage{wrapfig}

\definecolor{cvprblue}{rgb}{0.21,0.49,0.74}
\usepackage[pagebackref,breaklinks,colorlinks,allcolors=cvprblue]{hyperref}

\def\confName{CVPR}
\def\confYear{2027}

\title{
Think, Compress, Answer: \\
Reinforcement Learning for Multimodal Grounding via Reasoning Compression
}

\author{
\makebox[\textwidth][c]{%
    \normalsize
    Jinbae Seo$^{1}$ \quad
    Hyungjoon Kwon$^{1}$ \quad
    Jiyoung Lee$^{4}$ \quad
    Hyunsung Jang$^{1,3}$ \quad
    Namkoo Ha$^{3}$ \quad
    Jungin Park$^{2\dagger}$ \quad
    Kwanghoon Sohn$^{1\dagger}$
}
\\[0.8em]
\parbox{\textwidth}{%
    \centering\small
    $^{1}$Yonsei University \qquad
    $^{2}$DGIST \qquad
    $^{3}$LIG Defense \& Aerospace \\
    $^{4}$Division of Artificial Intelligence \& Software, Ewha Womans University
}
}

\begin{document}

\maketitle

\begin{abstract}
    Extended reasoning has substantially improved multimodal large language models (MLLMs) on tasks requiring compositional and relational inference. 
    However, multimodal grounding imposes a distinct requirement: reasoning must ultimately converge on a compact, target-specific representation that supports precise localization. 
    As reasoning trajectories grow longer, they often expand into broad scene narratives and peripheral relational details, which can dilute grounded perceptual evidence and lead to ambiguous or incorrect predictions. 
    Motivated by the information bottleneck principle, we propose Think-Compress-Answer (TCA), a reinforcement learning framework that decomposes multimodal grounding into three phases: free-form reasoning, target-aware context compression, and answer generation conditioned on the compressed context. 
    TCA is optimized through a multi-stage Group Relative Policy Optimization (GRPO) with progressive rewards that establish the structured output protocol, minimize reasoning-specific information while preserving target-relevant content, and directly improve localization accuracy. 
    This staged design enables MLLMs to retain the benefits of extended reasoning while distilling it into the minimal context required for grounding. 
    Experiments across six multimodal grounding benchmarks show that TCA consistently outperforms chain-of-thought and context-compression baselines, achieving zero-shot grounding performance comparable to or surpassing fully supervised methods.
\end{abstract}

\section{Introduction}
\label{sec:intro}
    \begin{figure}[t]
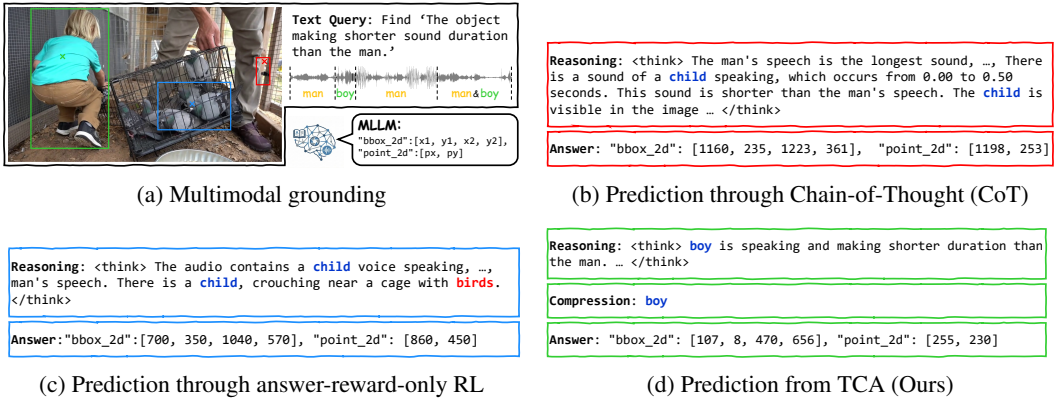

    \centering
        \begin{subfigure}[b]{0.49\textwidth}
            \centering
            \includegraphics[width=\linewidth]{Figure/fig1a.pdf}
            \caption{Multimodal grounding}
            \label{fig:1a}
        \end{subfigure}
        \\
        \begin{subfigure}[b]{0.49\textwidth}
            \centering
            \includegraphics[width=\linewidth]{Figure/fig1b.pdf}
            \caption{Prediction through Chain-of-Thought (CoT)}
            \label{fig:1b}
        \end{subfigure}
        \\        
        \begin{subfigure}[b]{0.49\textwidth}
            \centering
            \includegraphics[width=\linewidth]{Figure/fig1c.pdf}
            \caption{Prediction through answer-reward-only RL}
            \label{fig:1c}
        \end{subfigure}
        \\
        \begin{subfigure}[b]{0.49\textwidth}
            \centering
            \includegraphics[width=\linewidth]{Figure/fig1d.pdf}
            \caption{Prediction from TCA (Ours)}
            \label{fig:1d}
        \end{subfigure}
    
        \caption{
            \textbf{Motivation of \method{}.}
            Given multimodal inputs, (\red{b}) standard CoT and
            (\blue{c}) answer-reward-only RL methods identify the correct entity but drift during localization.
            (\green{d}) Our \method{} distills reasoning into a concise, faithful cue, yielding accurate grounding.
            The predicted bounding boxes and center points in (a) and the output from each method (b--d) are matched in color.
        }
        \label{fig:motivation}
        \vspace{-1em}
    \end{figure}

    \begin{figure}[t]
        \centering
        \includegraphics[width=\linewidth]{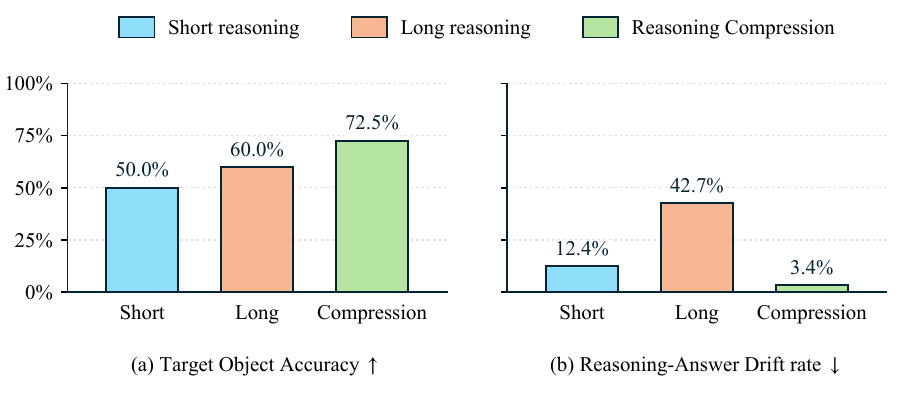}
    
        \caption{
            \textbf{Impact of reasoning length and compression on multimodal grounding.}
            Comparison of (a) target object identification accuracy and (b) reasoning--answer drift rate across short reasoning, long reasoning, and reasoning compression.
        }
        \label{fig:reasoning_tradeoff}
        \vspace{-1em}
    \end{figure}
    
    Multimodal grounding requires a model to identify the entity, event, or region in multimodal observations that is referred to by a text query.
    Although the final output is typically compact (\eg, bounding box, segmentation mask, center point, or temporal interval~\cite{refcoco, refytvos, refdavis17, mevis, refavs}), the model must interpret the query, distinguish among perceptually similar candidates, resolve spatial or temporal relations, and integrate evidence distributed across entities, observations, and modalities.
    Recent multimodal large language models (MLLMs) have supported these capabilities through explicit reasoning procedures, including chain-of-thought (CoT) prompting~\cite{wei2022chain, zhang2023multimodal}, instruction-following~\cite{llava, instructblip}, process supervision and intermediate verification~\cite{ddcot, visual-cot}, and test-time scaling~\cite{deepseekr1, openai_o1, snell2024scaling} that allocates additional computation to explore and refine reasoning paths.
    These advances suggest that reasoning is important for grounding targets that cannot be identified through direct query-perception matching alone.
    However, multimodal grounding also imposes a requirement that differs from open-ended reasoning: the reasoning process must ultimately isolate the compact, target-specific evidence necessary for precise localization.
    This raises a fundamental question: \textit{How can MLLMs retain the benefits of extended reasoning while preserving only the information required for grounding?}

    As reasoning paths grow longer, MLLMs often drift away from perceptual evidence, rely increasingly on language priors, and introduce hallucinated or peripheral details~\cite{zhou2024lure,pope,morethinking}.
    In multimodal grounding, extended CoT frequently expands target-oriented analysis into broad scene descriptions and relational narratives involving distractor objects.
    Moreover, conventional supervision—whether through CoT demonstrations or final task rewards in reinforcement learning (RL)—encourages accurate end predictions without specifying which intermediate evidence should be retained or discarded.
    Consequently, longer reasoning chains or additional test-time trajectories can identify the right target during reasoning yet still drift during coordinate prediction, as illustrated in \figref{fig:1b} and \figref{fig:1c}.


    To quantify this tension, we analyze how reasoning length affects target object identification versus reasoning--answer drift (\figref{fig:reasoning_tradeoff}).
    Extending reasoning improves target identification accuracy from 50.0\% to 60.0\%, but inflates the reasoning--answer drift rate from 12.4\% to 42.7\%.
    Simply restricting reasoning length curbs drift at the cost of weaker relational inference, while generic text summarization lacks grounding-aware supervision to filter out distractors.
    In contrast, selective reasoning compression achieves 72.5\% target identification accuracy with a drift rate of only 3.4\%.


    This trade-off naturally connects multimodal grounding to the information bottleneck (IB) principle~\cite{tishby2000information, tishby2015deep}, which seeks a minimal intermediate representation that discards non-predictive details while preserving task-relevant information.
    Viewed through the IB principle, existing reasoning-based MLLMs lack an explicit mechanism to compress verbose scene reasoning into the target-specific context needed for localization.
    Building on this insight, we propose \textbf{Think-Compress-Answer (TCA)}, an RL framework that decomposes multimodal grounding into three sequential phases: free-form reasoning, target-aware compression, and answer generation conditioned on the compressed context (\figref{fig:1d}).
    
    Because ground-truth labels supervise only final localization rather than intermediate compression, we optimize TCA through a four-stage Group Relative Policy Optimization (GRPO)~\cite{shao2024deepseekmath} curriculum that separates competing objectives.
    Stage~1 establishes the structured TCA output protocol via a format reward~\cite{deepseekr1}.
    Stage~2 minimizes a variational upper bound on the conditional mutual information between the reasoning path and compressed context, removing reasoning-specific verbosity.
    To prevent degenerate or overly generic compression, Stage~3 employs a specialized MLLM judge to verify that the compressed context remains concise, faithful to the reasoning, and aligned with the target.
    Finally, Stage~4 directly optimizes localization accuracy with a task-dependent grounding reward.

    We evaluate TCA across six benchmarks spanning referring expression segmentation (RES) and referring video object segmentation (R-VOS).
    Without modifying the MLLM or segmentation decoder architecture, TCA consistently outperforms standard CoT, answer-reward RL, and generic compression baselines, achieving zero-shot grounding performance comparable to or surpassing fully supervised methods.
    
    
    In summary, our contributions are threefold:
    \begin{itemize}
        \item We identify the trade-off between target identification and reasoning--answer drift in multimodal grounding, formulating selective reasoning compression through the information bottleneck (IB) principle.
        \item We propose Think-Compress-Answer (TCA), an RL framework that inserts an explicit target-aware compression bottleneck between reasoning and localization, optimized via a progressive four-stage GRPO curriculum.
        \item Extensive zero-shot experiments across six RES and R-VOS benchmarks show that TCA consistently outperforms CoT, answer-reward RL, and context-compression baselines while matching or exceeding fully supervised methods.
    \end{itemize}


\section{Related Work}
\label{sec:related}

\paragrapht{Chain-of-Thought (CoT) reasoning and multimodal grounding.}
    CoT prompting improves complex inference via intermediate reasoning steps~\cite{wei2022chain,kojima2022large}, with multimodal extensions introducing duty-distinct decomposition and spatial verification~\cite{zhang2023multimodal,ddcot,visual-cot,grounded-cot}.
    In multimodal grounding, reasoning is integrated either \emph{implicitly}, connecting MLLMs to decoders via latent tokens~\cite{lisa,pixellm,visa,villa}, or \emph{explicitly}, generating text chains before predicting coordinates or masks across visual and audio-visual inputs~\cite{cotrvs,meerkat,segzero,segr1,samr1,videosegr1,phasevos}.
    However, extended CoT chains often cause ``overthinking''~\cite{chen2024overthinking}, drifting into unfaithful steps~\cite{turpin2023unfaithful} or hallucinated details~\cite{huang2023hallucination}.
    Mitigations such as self-consistency~\cite{wang2023selfconsistency}, process rewards~\cite{lightman2023lets}, and test-time scaling~\cite{openai_o1,snell2024scaling,r1-vl} incur heavy compute while still exposing answer generation to verbose traces.
    Meanwhile, token-budgeting and context-compression methods~\cite{tale,acon} shorten intermediate text without target-specific grounding supervision, and DR$^2$Seg~\cite{drseg} checks description self-containment via two-rollout answer consistency.
    Unlike latent tokens, unconstrained CoT, or generic compression, TCA introduces a target-aware textual bottleneck that distills reasoning into a concise cue before localization.


\paragrapht{Reinforcement learning from AI feedback and GRPO.}
    While RLHF~\cite{ouyang2022training} relies on costly human annotations, RLAIF~\cite{lee2023rlaif,bai2022constitutional} and LLM-as-a-judge~\cite{zheng2023judging} provide scalable AI feedback.
    Group Relative Policy Optimization (GRPO)~\cite{shao2024deepseekmath} eliminates the critic via group-normalized advantages, and multi-stage RL~\cite{deepseekr1} stabilizes training through sequential objective decomposition.
    Recent grounding methods~\cite{segzero,segr1,samr1,videosegr1} apply GRPO with final localization rewards, supervising coordinates while leaving intermediate reasoning unconstrained.
    In contrast, we pair an MLLM-judge reward with a four-stage GRPO curriculum, dedicating each stage to a distinct requirement of bottleneck learning.


\paragrapht{Information bottleneck in deep learning.}
    The Information Bottleneck (IB) principle~\cite{tishby2000information} extracts compressed representations that preserve task-relevant content.
    \cite{alemi2017deep} instantiate IB in neural networks via variational bounds in continuous latent spaces, inspiring subsequent analyses of generalization~\cite{shwartz2017opening}.
    However, applying IB to \emph{discrete} natural language remains challenging because mutual information over token sequences is intractable.
    Rather than relying on continuous latents, we structure the model's output around an observable textual bottleneck, combining a tractable variational compression bound with an MLLM fidelity judge for GRPO training.


\section{Method}
\label{sec:method}

    \begin{figure*}[!t]
    \centering
    \includegraphics[width=\linewidth]{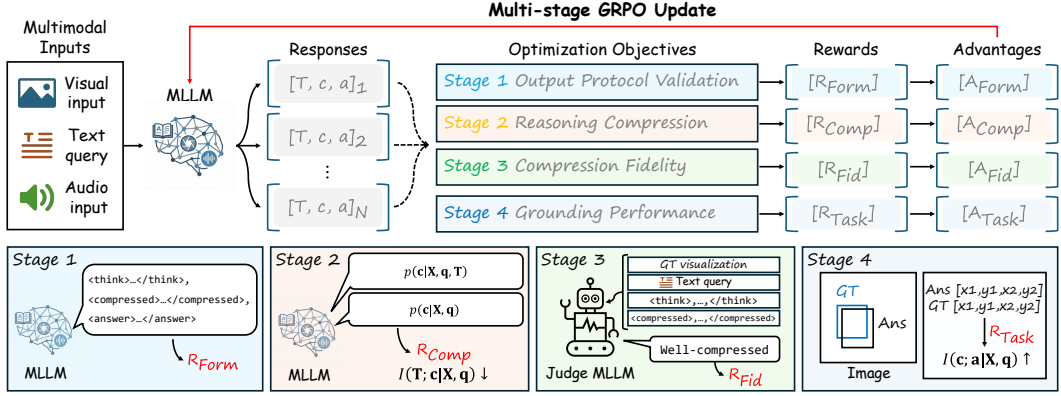}
    \caption{
    \textbf{Overview of \method{}.}
    The model generates $N$ grouped responses $[\mathbf{T}, \mathbf{c}, \mathbf{a}]$ per input.
    Stage~1 establishes the required output protocol, Stage~2 encourages reasoning compression, Stage~3 uses a judge MLLM to promote compression fidelity and target relevance, and Stage~4 improves grounding accuracy.
    The stage-specific rewards are converted into group-relative advantages to progressively update the model.}
    \label{fig:pipeline}
    \end{figure*}

\subsection{Motivation and overview}\label{sec:motivation}
    Given a text query $\mathbf{q}$ and another modality input $\mathbf{X}$ composed of observations from one or more modalities $\mathcal{M}$ (\eg, image, video, audio, or their combinations), multimodal grounding aims to identify the target entity, event, or region in $\mathbf{X}$ that is semantically referred to by $\mathbf{q}$:
    \begin{equation}
        \mathbf{y} = \mathcal{F}(\mathbf{X}, \mathbf{q}),
    \end{equation}
    where $\mathbf{y}\in \mathcal{Y}$ denotes the grounding target predicted by a grounding model $\mathcal{F}$ in a task-dependent output form $\mathcal{Y}$, such as a bounding box~\cite{referformer}, segmentation mask~\cite{lisa}, or center point~\cite{segzero}.

    Recent approaches~\cite{lisa,omnir1,segzero,revseg} have defined the grounding model $\mathcal{F}$ as a combination of a multimodal large language model (MLLM) $\pi$ and a decoder $\mathcal{D}$ that maps the output of $\pi$ into the target grounding space $\mathcal{Y}$.
    In existing reasoning-based MLLMs, inference typically follows an autoregressive reasoning-to-answer process:

    the model generates extended reasoning paths $\mathbf{T}=\{t_l\}_{l=1}^{L}$, where $L$ is the token length, produces an intermediate textual answer $\mathbf{a}$ (\ie, the target bounding-box and center-point coordinates) conditioned on $\mathbf{T}$, and then generates the final segmentation mask $\mathbf{y}$:
    \begin{equation}\label{eq:grounding}
        \mathbf{T} \sim \pi(\cdot \mid \mathbf{X}, \mathbf{q}), \quad \mathbf{a} \sim \pi(\cdot \mid \mathbf{X}, \mathbf{q}, \mathbf{T}), \quad \mathbf{y} = \mathcal{D}(\mathbf{a}, \mathbf{X}).
    \end{equation}

    With reinforcement learning (RL), such models are optimized to produce the reasoning paths $\mathbf{T}$ and the answer $\mathbf{a}$ that maximize task reward, encouraging $\mathbf{T}$ to retain information that is useful for predicting the intermediate target $\mathbf{a}$.
    In multimodal grounding, such reasoning remains important for identifying the correct grounding target.
    However, not all information contained in $\mathbf{T}$ is required for predicting $\mathbf{a}$.
    The key mismatch is that the intermediate target $\mathbf{a}$ is highly compact, whereas $\mathbf{T}$ often contains rich scene-level descriptions, relational analysis, and modality-wide contextual details that far exceed what is necessary for localization.
    
    To address this challenge, we propose TCA (Think-Compress-Answer) to decompose the reasoning-to-answer process.
    We design an intermediate reasoning compression stage to extract a target-specific compressed context $\mathbf{c}$ motivated by the information bottleneck (IB) principle~\cite{tishby2000information}.
    Instead of directly mapping the full reasoning paths $\mathbf{T}$ onto the answer (\ie, maximizing $I(\mathbf{T}; \mathbf{a})$), the model learns to retain only the information that is maximally predictive of $\mathbf{a}$, following the IB objective:
    \begin{equation}\label{eq:ib}
        \min_{\pi}
        \underbrace{
            I(\mathbf{T};\mathbf{c}\mid\mathbf{X},\mathbf{q})
        }_{\text{reasoning-specific information}}
        -
        \beta
        \underbrace{
            I(\mathbf{c};\mathbf{a}\mid\mathbf{X},\mathbf{q})
        }_{\text{task-relevant information}},
    \end{equation}
    where $\beta$ controls the trade-off between the amount of preserved information from the reasoning paths, $I(\mathbf{T}; \mathbf{c} \mid \mathbf{X}, \mathbf{q})$, and task-relevant information, $I(\mathbf{c}; \mathbf{a} \mid \mathbf{X}, \mathbf{q})$.
    The overall procedure of TCA is illustrated in \figref{fig:pipeline}.

\subsection{Think-Compress-Answer}\label{subsec:tca}
    TCA represents the extended reasoning paths $\mathbf{T}$, target-specific compressed context $\mathbf{c}$, and intermediate answer $\mathbf{a}$ as three explicit textual fields.
    In addition to \eqref{eq:grounding}, TCA introduces an intermediate process for $\mathbf{c}$ as
    \begin{equation}
        \mathbf{c} \sim \pi(\cdot\mid\mathbf{X},\mathbf{q},\mathbf{T}).
    \end{equation}

    The main challenge in training TCA is that no direct supervision is available for the optimal compressed context $\mathbf{c}$.
    Although $\mathbf{c}$ should be concise, derived from $\mathbf{T}$, and sufficient for predicting the correct grounding target, these properties are only indirectly constrained by the final grounding annotation $\mathbf{y}^*$.
    We therefore introduce four reward components, each of which targets a distinct property of TCA:
    a format reward that enforces a valid output structure;
    a compression-rate reward that removes reasoning-specific information from $\mathbf{c}$; 
    a compression-fidelity reward that preserves the target-relevant evidence required for grounding;
    and a grounding reward that directly evaluates localization accuracy.

    These rewards are optimized through a multi-stage GRPO curriculum~\cite{shao2024deepseekmath} to avoid jointly optimizing competing objectives from the beginning or shortcut solutions, such as bypassing the bottleneck by encoding information directly in the answer while leaving the compressed context uninformative~\cite{alemi2017deep}.
    At each training iteration, we sample $N$ responses from the current rollout policy $\bar{\pi}$, compute the rewards $R^n$ corresponding to each stage, and normalize the rewards to estimate a relative advantage for each response:
    \begin{equation}
        A^n = \frac{R^n - \text{mean}(\{R^1, \ldots, R^N\})}{\text{std}(\{R^1, \ldots, R^N\})}
    \end{equation}
    The model is then updated using the GRPO objective.

    \paragrapht{Stage 1 -- Format reward.} We first establish the desired output protocol using a simple format reward~\cite{deepseekr1}, encouraging the model to reliably separate free-form reasoning from the structured answer block.
    Let $\mathbf{o}$ denote the serialized output generated by $\pi$:
    \begingroup
    \footnotesize
    \begin{equation}
    \begin{aligned}
        \mathbf{o} = \bigl[
        &\texttt{<think>},\mathbf{T},\texttt{</think>},
        \\& \texttt{<compressed>}, \mathbf{c},\texttt{</compressed>},
        \\& \texttt{<answer>},\mathbf{a},\texttt{</answer>}
        \bigr].
    \end{aligned}
    \end{equation}
    \endgroup
    We define the format reward $R_\text{Form}$ as
    \begin{equation}\label{eq:rfmt}
    \begin{split}
        R_{\text{Form}}
        =
         \mathbf{1}[\texttt{<think>} \ldots \texttt{</think>} \in \mathbf{o}]
        \\ +
        \mathbf{1}[\texttt{<compressed>} \ldots \texttt{</compressed>} \in \mathbf{o}]
        \\ + 
        \mathbf{1}[\texttt{<answer>} \ldots \texttt{</answer>} \in \mathbf{o}]. 
    \end{split}
    \end{equation}
    The format reward yields a score in $\{0, 1, 2, 3\}$, normalized to $[0, 1]$, and provides partial credit for outputs that satisfy only a subset of the required constraints.

    \paragrapht{Stage 2 -- Compression-rate reward.}
    In Stage 2, we encourage $\mathbf{c}$ to discard reasoning-path-specific information by minimizing the first term in \eqref{eq:ib}.
    However, directly evaluating $I(\mathbf{T}; \mathbf{c} \mid \mathbf{X}, \mathbf{q})$ requires the intractable marginal $p(\mathbf{c}\mid\mathbf{X}, \mathbf{q})$.
    We therefore introduce a context-conditioned variational prior $\rho_\eta(\mathbf{c}\mid \mathbf{X}, \mathbf{q})$ that predicts the compressed context without observing $\mathbf{T}$.
    The conditional mutual information is upper-bounded~\cite{hou2025redundancy} as
    \begin{equation}\label{eq:rate_bound}
    \begin{split}
        I_{\pi}(\mathbf{T};\mathbf{c}\mid\mathbf{X},\mathbf{q}) \leq \mathbb{E} \big[ D_{\text{KL}}\big( \pi(\mathbf{c}\mid\mathbf{X},\mathbf{q},\mathbf{T}) \,\Vert\, \rho_{\eta}(\mathbf{c}\mid\mathbf{X},\mathbf{q}) \big) \big].
    \end{split}
    \end{equation}
    For a sampled trajectory, we estimate its information rate as
    \begin{equation}\label{eq:rate_estimator}
        \hat{\mathcal{I}}_\text{Rate} = \log \bar{\pi} (\mathbf{c}\mid \mathbf{X}, \mathbf{q}, \mathbf{T}) - \log \rho_\eta (\mathbf{c} \mid \mathbf{X}, \mathbf{q}),
    \end{equation}
    where $\bar{\pi}$ denotes the fixed rollout policy.
    For an autoregressive compressed context, each log-likelihood is computed by summing its token-level log-probabilities.
    The compression-rate reward is then defined as
    \begin{equation}\label{eq:rrate}
        R_\text{Comp} = \sigma(-\hat{\mathcal{I}}_\text{Rate}),
    \end{equation}
    where $\sigma(\cdot)$ is a sigmoid function to force the reward to be $[0, 1]$.
    Maximizing $R_\text{Comp}$ encourages $\mathbf{c}$ to contain less information that can only be recovered from the particular realization of $\mathbf{T}$.
    We instantiate the variational prior using the same rollout policy $\bar{\pi}$, \ie,
    $\rho_\eta (\mathbf{c} \mid \mathbf{X}, \mathbf{q}) := \bar{\pi}(\mathbf{c} \mid \mathbf{X}, \mathbf{q}).$
    The parameters of $\bar{\pi}$ are held fixed while computing rewards and performing each GRPO update, not requiring separately optimized variational prior.

    \paragrapht{Stage 3 -- Compression fidelity reward.}
    Minimizing the information rate alone admits a degenerate solution in which the model produces an empty or overly generic compressed context (\eg, ``object" or ``animal"), merely copies the input query, or generates an input-independent context~\cite{skalse2022rewardhacking}.
    We employ a specialized judge $\mathcal{J}$ (\ie, MLLM) to evaluate whether the compressed context $\mathbf{c}$ is well-aligned with $\mathbf{T}$, while preserving sufficient information for grounding.

    We first generate $\mathbf{X}^*$ by visualizing the ground-truth (\eg, segmentation mask, bounding box, or center point) on the input $\mathbf{X}$.
    The judge $\mathcal{J}$ takes $\mathbf{X}^*$, $\mathbf{q}$, $\mathbf{T}$, and $\mathbf{c}$ as inputs, and produces the fidelity reward as:    
    \begin{equation}
        R_\text{Fid} = \mathcal{J}(\mathbf{X}^*, \mathbf{q}, \mathbf{T}, \mathbf{c}),
    \end{equation}
    where $R_\text{Fid} \in \{0,1\}$ returns 1 if $\mathbf{c}$ is ``well-compressed" from $\mathbf{T}$ and semantically aligned with the ground-truth target highlighted on $\mathbf{X}^*$, and 0 otherwise.
    For more details on the prompt to generate the fidelity reward, please refer to the supplementary material.

    In Stage 2 and 3, GRPO updates are applied primarily to the tokens in $\mathbf{c}$, while the distribution of $\mathbf{T}$ is stabilized using a KL constraint to the Stage 1 checkpoint.
    This isolates compression learning from changes in the reasoning path distribution.

    \paragrapht{Stage 4 -- Grounding reward.}
    Finally, we optimize a task-driven surrogate for the task-relevant term $I(\mathbf{c};\mathbf{a}\mid\mathbf{X},\mathbf{q})$ in \eqref{eq:ib} using a reward on the intermediate answer and resulting grounding target.

    For segmentation-based grounding tasks, $\mathbf{a}$ consists of the predicted bounding-box coordinates $\mathbf{b}$ and the center point $\mathbf{p}$ of the target.
    From the ground-truth label $\mathbf{y}^*$, we construct the corresponding intermediate ground truth $\mathbf{a}^*$, consisting of the bounding box $\mathbf{b}^*$ and center point $\mathbf{p}^*$.
    We define the task reward as the sum of the bounding-box IoU, bounding-box $\ell_1$, and point $\ell_1$ rewards~\cite{segzero}:
    \begin{equation}
    \begin{split}
        R_\text{Task}
        & =
        \mathbf{1}[\text{IoU}(\mathbf{b},\mathbf{b}^*) > \tau_{\text{iou}}]
        \\& + \mathbf{1}[\lVert\mathbf{b}-\mathbf{b}^*\rVert_1 < \tau_b] +
        \mathbf{1}[\lVert\mathbf{p}-\mathbf{p}^*\rVert_1 < \tau_p].
    \end{split}
    \end{equation}
    The bounding-box IoU reward evaluates the overlap between $\mathbf{b}$ and $\mathbf{b}^*$, while the bounding-box and point $\ell_1$ rewards evaluate their distances from the corresponding ground truth.
    GRPO updates in Stage~4 are applied to both $\mathbf{c}$ and $\mathbf{a}$, while the reasoning field remains constrained by the previous-stage policy.
    We set $\tau_{\text{iou}}{=}0.5$, $\tau_b{=}10$, and $\tau_p{=}100$ in our experiments.
    
    Overall, the four-stage optimization first establishes a valid TCA protocol, then learns a minimum sufficient compressed context, and finally improves grounding accuracy without allowing the compressed context to recover the full information content of the original reasoning paths.

\section{Experiments}
\label{sec:experiments}

\subsection{Experimental Setup}
\label{subsec:setup}

\begin{table*}[!t]
\centering
\small
\caption{Referring expression segmentation on RefCOCO, RefCOCO+, and RefCOCOg.
We report the performance (cIoU) evaluated on validation, testA, and testB splits for RefCOCO and RefCOCO+, and validation and test splits for RefCOCOg.
* employs a frozen decoder (Dec.) only for mask prediction.}
\label{tab:refcoco_main}

\begin{tabular}{lllcccccccc}
\toprule
\multirow{2}{*}{Method}
& \multirow{2}{*}{MLLM}
& \multirow{2}{*}{Dec.}
& \multicolumn{3}{c}{RefCOCO}
& \multicolumn{3}{c}{RefCOCO+}
& \multicolumn{2}{c}{RefCOCOg} \\
\cmidrule(lr){4-6}
\cmidrule(lr){7-9}
\cmidrule(lr){10-11}
& & & val & tA & tB & val & tA & tB & val & test \\
\midrule

\multicolumn{11}{l}{\hspace{-6pt}\gray{\textit{Supervised, Fine-Tuning}}} \\
Qwen2.5-Omni-7B~\cite{qwen2.5omni}
& Qw2.5OM-7B & SAM2$^*$
& 76.2 & 78.0 & 71.3
& 69.4 & 75.1 & 62.7
& 71.5 & 70.5 \\

\multicolumn{11}{l}{\hspace{-6pt}\gray{\textit{Supervised, Embedding-Guided}}} \\
LISA~\cite{lisa}
& LLaVA-7B & SAM
& 74.1 & 76.5 & 71.1
& 62.4 & 67.4 & 56.5
& 66.4 & 68.5 \\

PixelLM~\cite{pixellm}
& LLaVA-7B & Pix.Dec.
& 73.0 & 76.5 & 68.2
& 66.3 & 71.7 & 58.3
& 69.3 & 70.5 \\

GSVA~\cite{gsva}
& Vicuna-7B & SAM
& 76.4 & 77.4 & 72.8
& 64.5 & 67.7 & 58.6
& 71.1 & 72.0 \\

GLaMM~\cite{glamm}
& Vicuna-7B & SAM
& 79.5 & \textbf{83.2} & 76.9
& 72.6 & 78.7 & 64.6
& 74.2 & 74.9 \\

OMG-LLaVA~\cite{omgllava}
& IntLM2-7B & OMG-Seg
& 78.0 & 80.3 & 74.1
& 69.1 & 73.1 & 63.0
& 72.9 & 72.9 \\

F-LMM~\cite{flmm}
& LLaVA-7B & SAM
& 75.2 & 79.1 & 71.9
& 63.7 & 71.8 & 54.7
& 67.1 & 68.1 \\

Sa2VA~\cite{sa2va}
& IntVL2.0-8B & SAM2
& \textbf{81.9} & -- & --
& \textbf{76.5} & -- & --
& \textbf{78.9} & -- \\

UniPixel~\cite{unipixel}
& Qw2.5VL-7B & SAM2
& 80.8 & 83.0 & \textbf{77.4}
& 75.3 & \textbf{80.1} & \textbf{70.0}
& 76.4 & \textbf{77.1} \\

\multicolumn{11}{l}{\hspace{-6pt}\gray{\textit{Supervised, Reasoning-Guided}}} \\
SAM4MLLM~\cite{sam4mllm}
& LLaVA1.6-7B & SAM$^*$
& 79.6 & 82.8 & 76.1
& 73.5 & 77.8 & 65.8
& 74.5 & 75.6 \\

Seg-Zero~\cite{segzero}
& Qw2.5VL-7B & SAM2$^*$
& -- & 80.3 & --
& -- & 76.2 & --
& -- & 72.6 \\

DR$^2$Seg-7B~\cite{drseg}
& Qw2.5VL-7B & SAM2$^*$
& -- & 78.7 & --
& -- & 75.4 & --
& -- & 72.2 \\

SAM-R1~\cite{samr1}
& Qw2.5VL-7B & SAM2$^*$
& -- & 79.2 & --
& -- & 74.7 & --
& -- & 73.1 \\

\midrule
\multicolumn{11}{l}{\hspace{-6pt}\gray{\textit{Zero-Shot, Reasoning-Guided}}} \\
Qwen2.5-Omni-7B~\cite{qwen2.5omni}
& Qw2.5OM-7B & SAM2$^*$
& 74.5 & 75.8 & 69.1
& 66.5 & 71.2 & 60.5
& 71.8 & 71.2 \\

TALE~\cite{tale}
& Qw2.5OM-7B & SAM2$^*$
& 75.4 & 76.7 & 70.7
& 67.5 & 72.9 & 60.3
& 73.0 & 72.3 \\

ACON~\cite{acon}
& Qw2.5OM-7B & SAM2$^*$
& 74.3 & 75.9 & 69.7
& 67.0 & 70.7 & 60.1
& 72.5 & 71.5 \\

Seg-R1-7B~\cite{segr1}
& Qw2.5VL-7B & SAM2$^*$
& 74.3 & 78.7 & 67.6
& 62.6 & 70.9 & 57.9
& 71.0 & 71.4 \\

VideoSeg-R1~\cite{videosegr1}
& Qw2.5VL-7B & SAM2$^*$
& 78.2 & \textbf{82.3} & \textbf{75.1}
& 71.8 & 76.1 & 64.7
& 73.1 & 74.1 \\

\methodshort{} (\textbf{Ours})
& Qw2.5OM-7B & SAM2$^*$
& \textbf{79.4} & 81.5 & 73.6
& \textbf{74.0} & \textbf{78.7} & \textbf{66.1}
& \textbf{75.3} & \textbf{76.0} \\

\bottomrule
\end{tabular}
\end{table*}

\paragrapht{Datasets.}
    We primarily evaluate TCA in a zero-shot setting across six benchmarks spanning two tasks: RefCOCO, RefCOCO+, and RefCOCOg~\cite{refcoco,refcocog} for referring expression segmentation (RES), and Ref-YouTube-VOS (RY-V)~\cite{refytvos}, Ref-DAVIS17 (RD17)~\cite{refdavis17}, and MeViS~\cite{mevis} for referring video object segmentation (R-VOS).
    TCA is trained for one epoch on a mixed set of 2,100 samples comprising 887 audio-visual referring expressions from RefAVS~\cite{refavs} and 1,213 reasoning video expressions from ReVOS~\cite{visa}.
    Neither RES nor R-VOS data is used during training.
    We also evaluate in-domain reasoning VOS and audio-visual segmentation on ReVOS and RefAVS, with full dataset statistics provided in the supplementary material.

\paragrapht{Baselines.}
    We compare TCA against two groups of MLLM grounding approaches categorized by how $\pi$ interfaces with the mask decoder $\mathcal{D}$:
    (1) \textit{Embedding-guided methods}~\cite{lisa,pixellm,glamm,sa2va,unipixel}, which pass latent grounding tokens from $\pi$ to $\mathcal{D}$ and jointly train both modules on target-domain masks; and
    (2) \textit{Reasoning-guided methods}~\cite{segzero,samr1,drseg,segr1,videosegr1,phasevos,omnir1}, which pair $\pi$ with an off-the-shelf frozen decoder (\eg, SAM2~\cite{sam2}) conditioned on explicit coordinate predictions $\mathbf{a}$.
    TCA follows the reasoning-guided paradigm.
    For controlled comparisons under the same Qwen2.5-Omni-7B~\cite{qwen2.5omni} backbone and frozen SAM2 decoder, we evaluate zero-shot CoT prompting, supervised fine-tuning (SFT), and two LLM reasoning summarization and context-compression baselines: TALE~\cite{tale} and ACON~\cite{acon}.

\paragrapht{Backbone and implementation.}
    We instantiate $\pi$ with Qwen2.5-Omni-7B~\cite{qwen2.5omni} and $\mathcal{D}$ with frozen SAM2~\cite{sam2}, where $\pi$ predicts bounding-box and center-point coordinates $\mathbf{a}$ to prompt $\mathcal{D}$ for the segmentation mask $\mathbf{y}$.
    We apply LoRA~\cite{lora} only to the LLM layers of $\pi$ while keeping all pre-trained parameters in $\pi$ and $\mathcal{D}$ frozen.
    Additional implementation details and hyperparameters are reported in the supplementary material.

\subsection{Referring Expression Segmentation}
\label{subsec:res}
    Given an image and a natural language referring expression, RES aims to generate a pixel-wise mask for the referred target (\tabref{tab:refcoco_main}).
    Under the identical Qwen2.5-Omni-7B backbone and frozen SAM2 decoder, TCA improves over zero-shot CoT across all eight splits, gaining +4.9, +7.5, and +4.1 cIoU on the validation sets of RefCOCO, RefCOCO+, and RefCOCOg, while consistently outperforming SFT (\eg, +4.6 cIoU on RefCOCO+ val and +4.8 on RefCOCOg test).
    Generic compression baselines (TALE~\cite{tale} and ACON~\cite{acon}) bring only minor gains on RefCOCO and slightly degrade RefCOCO+ testB, whereas TCA surpasses TALE by +6.5 cIoU on appearance-centric RefCOCO+ val, where location terms are disallowed and multi-object reasoning is prone to drift.

    Compared with reasoning-guided RL baselines, TCA outperforms zero-shot Seg-R1-7B~\cite{segr1} across all splits, exceeds VideoSeg-R1~\cite{videosegr1} on six of eight splits (including every split of RefCOCO+ and RefCOCOg), and improves over supervised Seg-Zero~\cite{segzero} by +1.2, +2.5, and +3.4 cIoU on RefCOCO testA, RefCOCO+ testA, and RefCOCOg test, respectively.
    Despite zero-shot evaluation with a frozen decoder, TCA also matches or surpasses most end-to-end supervised embedding-guided models~\cite{lisa,pixellm,glamm,omgllava}.

    \begin{table*}[!t]
\centering
\small
\caption{Referring video object segmentation on Ref-YouTube-VOS, Ref-DAVIS17, and MeViS.
We report $\mathcal{J}\&\mathcal{F}$ evaluated on the validation set of each benchmark.
* employs a frozen decoder (Dec.) only for mask prediction.
$\dagger$ additionally employs an external XMem tracker for temporal mask propagation.}
\label{tab:rvos_main}

\begin{tabular}{lllccc}
\toprule
Method
& MLLM
& Dec.
& RY-VOS
& RD17
& MeViS \\
\midrule

\multicolumn{6}{l}{\hspace{-6pt}\gray{\textit{Supervised, Fine-Tuning}}} \\
Qwen2.5-Omni-7B~\cite{qwen2.5omni}
& Qw2.5OM-7B
& SAM2$^*$
& 69.3
& 76.8
& 53.3 \\

\multicolumn{6}{l}{\hspace{-6pt}\gray{\textit{Supervised, Embedding-Guided}}} \\
VISA~\cite{visa}
& ChatUniVi-7B
& SAM$^\dagger$
& 61.5
& 69.4
& 43.5 \\

ViLLa~\cite{villa}
& Vicuna-7B
& E2E
& 67.5
& 74.3
& 49.4 \\

Sa2VA~\cite{sa2va}
& IntVL2.0-8B
& SAM2
& 70.7
& 75.2
& 46.9 \\

VideoLISA~\cite{videolisa}
& LLaVA-Phi-3.8B
& SAM$^\dagger$
& 63.7
& 68.8
& 44.4 \\

UniPixel~\cite{unipixel}
& Qw2.5VL-7B
& SAM2
& 71.0
& 76.4
& \textbf{55.8} \\

\multicolumn{6}{l}{\hspace{-6pt}\gray{\textit{Supervised, Reasoning-Guided}}} \\
VideoSeg-R1~\cite{videosegr1}
& Qw2.5VL-7B
& SAM2$^{*\dagger}$
& \textbf{81.3}
& \textbf{79.8}
& 55.3 \\

\midrule
\multicolumn{6}{l}{\hspace{-6pt}\gray{\textit{Zero-Shot, Reasoning-Guided}}} \\

Qwen2.5-Omni-7B~\cite{qwen2.5omni}
& Qw2.5OM-7B
& SAM2$^*$
& 63.5
& 65.2
& 49.1 \\

TALE~\cite{tale}
& Qw2.5OM-7B
& SAM2$^*$
& 67.0
& 71.9
& 50.4 \\

ACON~\cite{acon}
& Qw2.5OM-7B
& SAM2$^*$
& 66.6
& 70.7
& 50.8 \\

PARSE-VOS~\cite{phasevos}
& Llama3-8B
& SAM2$^*$
& 72.1
& 75.5
& 52.4 \\

Omni-R1~\cite{omnir1}
& Qw2.5OM-7B
& SAM2$^*$
& 71.0
& 77.6
& 52.3 \\

\methodshort{} (\textbf{Ours})
& Qw2.5OM-7B
& SAM2$^*$
& \textbf{72.3}
& \textbf{78.9}
& \textbf{55.8} \\

\bottomrule
\end{tabular}
\end{table*}

\begin{figure*}[!t]
    \centering

    \begin{minipage}[c]{0.4\linewidth}
        \centering

        \captionof{table}{
            Stage-wise ablations of TCA under an equal training budget on R-VOS validation sets ($\mathcal{J}\&\mathcal{F}$).
        }
        \label{tab:rvos_ablation}

        \small
        \setlength{\tabcolsep}{3.5pt}

        \begin{tabular}{ccccccc}
        \toprule
        S1 & S2 & S3 & S4 & RY-V & RD17 & MeViS \\
        \midrule
        $\times$ & $\times$ & $\times$ & $\times$ & 63.5 & 65.2 & 49.1 \\
        $\checkmark$ & $\times$ & $\times$ & $\times$ & 65.6 & 69.6 & 50.6 \\
        $\checkmark$ & $\checkmark$ & $\times$ & $\times$ & 69.0 & 72.6 & 52.3 \\
        $\checkmark$ & $\checkmark$ & $\checkmark$ & $\times$ & 71.1 & 75.3 & 53.0 \\
        \midrule
        $\checkmark$ & $\times$ & $\times$ & $\checkmark$ & 68.6 & 72.8 & 51.5 \\
        $\checkmark$ & $\checkmark$ & $\times$ & $\checkmark$ & 70.8 & 76.1 & 53.8 \\
        \midrule
        $\checkmark$ & $\checkmark$ & $\checkmark$ & $\checkmark$
        & \textbf{72.3} & \textbf{78.9} & \textbf{55.8} \\
        \bottomrule
        \end{tabular}

    \end{minipage}
    \hfill
    \begin{minipage}[c]{0.59\linewidth}
        \centering
        \vspace{1em}

        \begin{subfigure}[b]{0.62\linewidth}
            \centering
            \includegraphics[width=\linewidth]{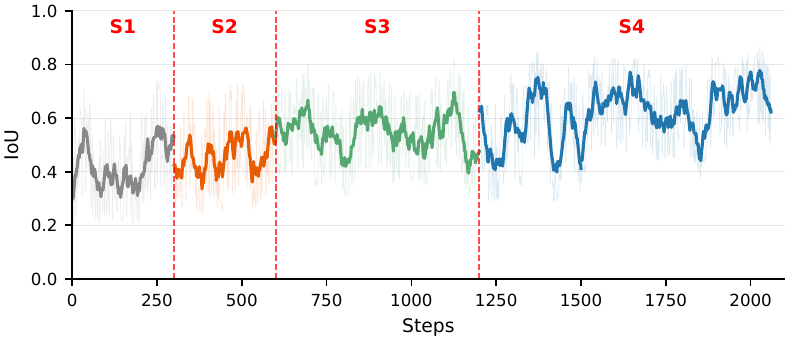}
            \caption{Grounding IoU}
            \label{fig:stage_grounding}
        \end{subfigure}
        \hfill
        \begin{subfigure}[b]{0.35\linewidth}
            \centering
            \includegraphics[width=\linewidth]{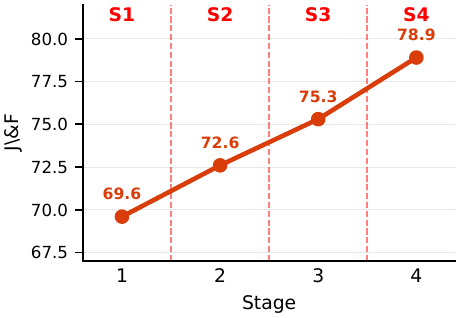}
            \caption{Zero-shot $\mathcal{J}\&\mathcal{F}$}
            \label{fig:stage_perform}
        \end{subfigure}

        \captionof{figure}{
            \textbf{Training and stage progression.}
            (a) Grounding IoU over training steps across Stages~1--4.
            (b) Zero-shot $\mathcal{J}\&\mathcal{F}$ on Ref-DAVIS17 after each stage.
        }
        \label{fig:stage}

    \end{minipage}
\end{figure*}

\subsection{Referring Video Object Segmentation}
\label{subsec:r-vos}
    R-VOS segments a referred object across video frames under motion, occlusion, and temporal appearance changes (\tabref{tab:rvos_main}).
    Evaluated zero-shot, TCA achieves 72.3, 78.9, and 55.8 $\mathcal{J}\&\mathcal{F}$ on RY-V, RD17, and MeViS, establishing state-of-the-art zero-shot performance while surpassing supervised embedding-guided approaches~\cite{visa,villa,sa2va,videolisa,unipixel} on RY-V and RD17 and tying UniPixel~\cite{unipixel} for the highest score on motion-focused MeViS.
    Under the same Qwen2.5-Omni-7B backbone, TCA improves upon zero-shot CoT by +8.8, +13.7, and +6.7 points, outperforms SFT by +3.0, +2.1, and +2.5 points, and exceeds Omni-R1~\cite{omnir1} across all three benchmarks (+3.5 on MeViS).
    Although TALE~\cite{tale} and ACON~\cite{acon} consistently improve over zero-shot CoT, TCA outperforms both by +5.0 to +8.2 $\mathcal{J}\&\mathcal{F}$, demonstrating the necessity of preserving target-specific perceptual cues through our IB-guided objective.

    \begin{figure*}[!t]
        \centering
        \includegraphics[width=\linewidth]{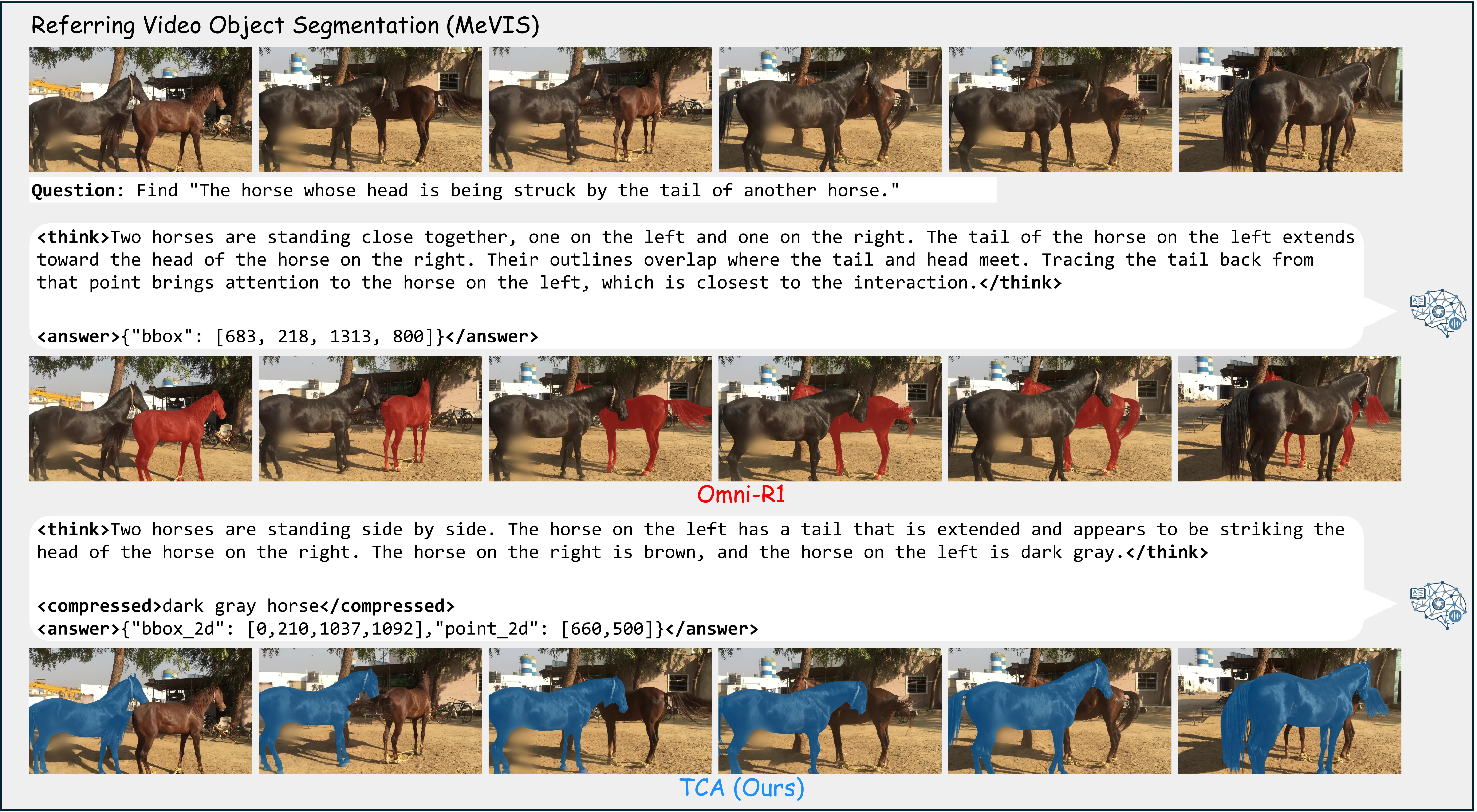}
        \caption{\textbf{Qualitative comparison on MeViS.} While Omni-R1 drifts across visually similar instances, TCA distills extended motion-appearance reasoning into the concise cue \texttt{dark gray horse}, yielding temporally consistent masks.}
        \label{fig:qualitative_mevis}
    \end{figure*}
    
\subsection{Ablation Study and Analysis}
\label{subsec:analysis}

\begin{table*}[t]
\centering
\small
\caption{Reasoning video object segmentation on the ReVOS~\cite{visa} benchmark and referring audio-visual segmentation on the RefAVS~\cite{refavs} benchmark.
We report region similarity ($\mathcal{J}$), contour accuracy ($\mathcal{F}$), and their average ($\mathcal{J}\&\mathcal{F}$). We use SAM2 as the decoder. 
}
\label{tab:revos_refavs}

\setlength{\tabcolsep}{4pt}
\resizebox{\linewidth}{!}{%
\begin{tabular}{lll ccc ccc ccc}
\toprule
\multirow{2}{*}{Method}
& \multirow{2}{*}{MLLM}
& \multirow{2}{*}{Dec.}
& \multicolumn{3}{c}{ReVOS}
& \multicolumn{3}{c}{RefAVS (Seen)}
& \multicolumn{3}{c}{RefAVS (Unseen)} \\
\cmidrule(lr){4-6}
\cmidrule(lr){7-9}
\cmidrule(lr){10-12}
& & &
$\mathcal{J}\&\mathcal{F}$ & $\mathcal{J}$ & $\mathcal{F}$
& $\mathcal{J}\&\mathcal{F}$ & $\mathcal{J}$ & $\mathcal{F}$
& $\mathcal{J}\&\mathcal{F}$ & $\mathcal{J}$ & $\mathcal{F}$ \\
\midrule



\multicolumn{12}{l}{\hspace{-6pt}\gray{\textit{Zero-Shot, Reasoning-Guided}}} \\
Qwen2.5-Omni-7B~\cite{qwen2.5omni}
& Qw2.5OM-7B & SAM2
& 38.4 & 37.2 & 39.6
& 35.5 & 30.1 & 40.9
& 64.8 & 64.8 & 64.9 \\

TALE~\cite{tale}
& Qw2.5OM-7B & SAM2
& 49.7 & 45.0 & 54.4
& 56.1 & 53.7 & 58.5
& 67.7 & 67.7 & 67.8 \\

ACON~\cite{acon}
& Qw2.5OM-7B & SAM2
& 50.6 & 46.1 & 55.1
& 56.0 & 53.7 & 58.3
& 68.8 & 68.8 & 68.8 \\

\multicolumn{12}{l}{\hspace{-6pt}\gray{\textit{Supervised, Fine Tuning}}} \\
Qwen2.5-Omni-7B~\cite{qwen2.5omni}
& Qw2.5OM-7B & SAM2
& \underline{55.4} & \underline{53.0} & \underline{57.8}
& \underline{69.9} & \underline{67.7} & \underline{72.0}
& \underline{77.0} & \underline{77.0} & \underline{77.1} \\
\midrule




\multicolumn{12}{l}{\hspace{-6pt}\gray{\textit{Supervised, Reasoning-Guided}}} \\
Omni-R1~\cite{omnir1}
& Qw2.5OM-7B & SAM2
& 47.6 & -- & --
& 47.2 & 43.0 & 51.4
& 74.2 & 71.3 & 77.0 \\

\methodshort{} (\textbf{Ours})
& Qw2.5OM-7B & SAM2
& \textbf{60.5} & \textbf{57.7} & \textbf{63.2}
& \textbf{70.7} & \textbf{68.0} & \textbf{73.4}
& \textbf{78.8} & \textbf{78.7} & \textbf{78.8} \\
\bottomrule
\end{tabular}%
}
\end{table*}


\paragrapht{Reasoning VOS and audio-visual segmentation.}
    \tabref{tab:revos_refavs} evaluates TCA on the validation split of ReVOS~\cite{visa} and the Seen (RefAVS-S) and Unseen (RefAVS-U) test splits of RefAVS~\cite{refavs}.
    TCA achieves 60.5, 70.7, and 78.8 $\mathcal{J}\&\mathcal{F}$, improving over zero-shot Qwen2.5-Omni-7B by +22.1, +35.2, and +14.0 points, surpassing SFT across all splits (+5.1 on ReVOS and +1.8 on RefAVS-U), and outperforming Omni-R1~\cite{omnir1} by +12.9 to +23.5 points on ReVOS and RefAVS-S.

\paragrapht{Optimization stages.}
    \tabref{tab:rvos_ablation} ablates the four-stage curriculum under an equal step budget on R-VOS.
    Enforcing the output format in Stage~1 improves over the base model across all datasets, and Stage~2 rate reduction adds +3.4, +3.0, and +1.7 $\mathcal{J}\&\mathcal{F}$ on RY-V, RD17, and MeViS.
    Adding the Stage~3 fidelity judge reaches 71.1, 75.3, and 53.0 $\mathcal{J}\&\mathcal{F}$ \textit{before any localization reward is applied}, outperforming answer-reward-only GRPO without compression (S1+S4: 68.6, 72.8, and 51.5).
    Skipping Stage~3 fidelity verification (S1+S2+S4) lowers scores by 1.5--2.8 points relative to full TCA (72.3, 78.9, and 55.8), confirming that rate minimization and semantic fidelity are complementary.
    Consistent with this trend, \figref{fig:stage} shows that training grounding IoU rises steadily during Stages~2 and 3 prior to Stage~4 coordinate supervision, yielding monotonic zero-shot gains on RD17.

\paragrapht{Backbone, decoder, and judge generality.}
    TCA transfers effectively across model scales, decoders, and Stage~3 judges (detailed tables in the supplementary material).

\paragrapht{Qualitative results.}
    \figref{fig:qualitative_mevis} compares TCA with Omni-R1~\cite{omnir1} on MeViS.
    While both models identify relevant scene elements during reasoning, TCA isolates the distinguishing cue \texttt{dark gray horse} in its compressed context to anchor localization across frames.
    Compressed token-length distributions, additional visualizations, and failure cases are provided in the supplementary material.

\section{Conclusion}
\label{sec:conclusion}
    We presented Think-Compress-Answer (TCA), a novel information-bottleneck-inspired RL framework that makes explicit reasoning compatible with multimodal grounding.
    We optimize TCA with a staged GRPO using practical rewards, progressively enforcing a valid output protocol, compression rate, compression fidelity, and end-task grounding performance.
    Without architectural modification or dense intermediate annotations, TCA achieves comparable and even better zero-shot performance than the supervised methods on RES and R-VOS benchmarks.
    
    \paragrapht{Broader impact.} 
    TCA offers a practical framework to study the interaction between long-form reasoning and multimodal grounding.
    By explicitly optimizing compression via staged RL, TCA enables the community to (1) diagnose when extended reasoning becomes counterproductive, (2) measure information retained in the compressed context, and (3) align inference with compact grounding objectives.
    This perspective shifts the focus from ``more reasoning'' to ``selective reasoning,'' informing future work on test-time scaling, information bottlenecks, and multimodal evaluation.
    The release of our training recipes is expected to facilitate ablations on reward design, judge selection, and compression capacity, as well as standardized comparisons across datasets and modalities.


\section*{LLM Usage Statement}
We used OpenAI ChatGPT to assist with experimental code development, LaTeX formatting, and language editing of the manuscript.
All AI-assisted code and manuscript content were reviewed and verified by the authors.
We take full responsibility for the final content of this work, including all experimental results, text, and artifacts.

{
    \small
    \bibliographystyle{ieeenat_fullname}
    \bibliography{main}
}

\clearpage

\twocolumn[
    \begin{center}
        \LARGE\bfseries
        Supplementary Material
    \end{center}
    \vspace{1em}
]

\appendix

\makeatletter
\@addtoreset{table}{section}
\@addtoreset{figure}{section}
\makeatother

\renewcommand{\thetable}{\thesection\arabic{table}}
\renewcommand{\thefigure}{\thesection\arabic{figure}}

\renewcommand{\theHtable}{appendix.\thesection.\arabic{table}}
\renewcommand{\theHfigure}{appendix.\thesection.\arabic{figure}}

\section{Experimental Details}
\subsection{Algorithm}
    The overall procedure of our TCA is described in \algref{alg:tca_grpo}.
    \begin{algorithm}[h]
\caption{Stage-wise GRPO for Think-Compress-Answer}
\label{alg:tca_grpo}
\small
\begin{algorithmic}[1]
\Require Dataset $\mathcal{D}=\{(\mathbf{X}_i,\mathbf{q}_i,\mathbf{y}_i^*)\}$, policy $\pi$, judge MLLM $\mathcal{J}$, group size $N$, stage steps $\{K_s\}_{s=1}^{4}$
\Ensure Optimized policy $\pi^*$

\For{$s=1$ to $4$}
    \For{$k=1$ to $K_s$}
        \State Set rollout policy $\bar{\pi}\gets\pi$
        \State Sample $(\mathbf{X},\mathbf{q},\mathbf{y}^*)\sim\mathcal{D}$

        \For{$n=1$ to $N$}
            \State Sample $\mathbf{o}_n\sim\bar{\pi}(\cdot\mid\mathbf{X},\mathbf{q})$
            \State Parse $\mathbf{o}_n$ into $(\mathbf{T}_n,\mathbf{c}_n,\mathbf{a}_n)$

            \If{$s=1$}
                \State $r_n\gets R_{\mathrm{Form}}(\mathbf{o}_n)$
            \ElsIf{$s=2$}
                \State $r_n\gets R_{\mathrm{Comp}}(\mathbf{T}_n,\mathbf{c}_n)$
            \ElsIf{$s=3$}
                \State $\mathbf{X}^*\gets\operatorname{VisualizeGT}(\mathbf{X},\mathbf{y}^*)$
                \State $r_n\gets\mathcal{J}(\mathbf{X}^*,\mathbf{q},\mathbf{T}_n,\mathbf{c}_n)$
            \Else
                \State $r_n\gets R_{\mathrm{Task}}(\mathbf{a}_n,\mathbf{y}^*)$
            \EndIf
        \EndFor

        \For{$n=1$ to $N$}
            \State $A_n\gets
            \dfrac{r_n-\operatorname{mean}(\{r_j\}_{j=1}^{N})}
            {\operatorname{std}(\{r_j\}_{j=1}^{N})}$
        \EndFor

        \State Update $\pi$ using GRPO with $\{(\mathbf{o}_n,A_n)\}_{n=1}^{N}$
    \EndFor
\EndFor

\State \Return $\pi^*\gets\pi$
\end{algorithmic}
\end{algorithm}

\subsection{Datasets}\label{sec:dataset}
\subsubsection{Referring expression segmentation}
    We evaluate our TCA on three RES benchmarks built on MSCOCO~\cite{mscoco}:
\begin{itemize}
    \item \textbf{RefCOCO}~\cite{refcoco} is collected in an interactive game-based setting and contains 142k referring expressions for 50k objects in 19,994 images.
    It provides 10,834, 5,657, and 5,095 expressions for validation, testA, and testB splits, where testA and testB contain person-only and non-person objects, respectively.
    \item \textbf{RefCOCO+}~\cite{refcoco} has a similar scale with 142k expressions for 50k objects in 19,992 images, but prohibits location words during annotation, making the expressions more appearance-centric.
    Validation, testA, and testB splits include 10,758, 5,726, and 4,889 expressions, respectively.
    \item \textbf{RefCOCOg}~\cite{refcocog} provides longer and more descriptive expressions, with 95,010 expressions for 49,822 objects in 25,799 images, of which 4,896 and 9,602 expressions are in the validation and test splits of the UMD partition.
\end{itemize}

\subsubsection{Referring video object segmentation}
    We evaluate our TCA on three R-VOS benchmarks:
\begin{itemize}
    \item \textbf{Ref-YouTube-VOS} (RY-V)~\cite{refytvos} is a large-scale dataset built upon YouTube-VOS, comprising 3,978 videos with 15k referring expressions.
    The dataset contains 202 validation videos and 305 test videos.
    \item \textbf{Ref-DAVIS17} (RD17)~\cite{refdavis17} extends the densely annotated DAVIS 2017~\cite{davis2017} benchmark with language annotations, providing 807 referring expressions for 205 objects across 90 videos.
    The validation split includes 30 videos with 243 expressions for 61 objects.
    \item \textbf{MeViS}~\cite{mevis} focuses on motion-expression-guided segmentation, where expressions must describe objects through their motion patterns rather than static appearance.
    It contains 2,006 videos with 28,570 expressions, and the validation split provides 140 videos.
\end{itemize}
    Note that we \textit{do not} use the R-VOS benchmarks during training, but only for zero-shot performance evaluation.

\subsubsection{Reasoning video object segmentation}
We evaluate on ReVOS~\cite{visa}, which aggregates videos from LV-VIS~\cite{lvvis}, MOSE~\cite{mose}, MeViS~\cite{mevis}, OVIS~\cite{ovis}, TAO, and UVO~\cite{uvo} to form a large-scale reasoning VOS benchmark.
It contains 1,042 videos with 35k expressions, including implicit reasoning-oriented and explicit referring expressions, for 5,535 objects, where reasoning-oriented queries require inferring the target through actions, relations, or world knowledge rather than direct appearance descriptions.

The benchmark is split into 626 training videos and 416 validation videos.
We sample 1,213 expressions from 473 videos for training.
The list of sampled data will be provided in the code.

\subsubsection{Referring audio-visual segmentation}
We evaluate on Ref-AVS~\cite{refavs}, which provides 4,002 videos with 20,261 referring expressions that require joint reasoning over visual, audio, and language cues to segment the referred object.
It provides 2,288 and 1,454 expressions for seen and unseen test splits.
We use 887 expressions from 557 source videos for training.

\begin{table}[h]
\centering
\small
\caption{Hyperparameters for training} 
\label{tab:hyperparams}
\begin{tabular}{ll}
\toprule
Hyperparameter & Value \\
\midrule
Framework & ms-swift \\
Precision & bfloat16 \\
Attention & FlashAttention \\
Learning rate & $1\times10^{-6}$ \\
Optimizer & AdamW \\
LR scheduler & Cosine \\
Warmup ratio & 0.05 \\
Per-device train batch size & 8 \\
Generation batch size & 8 \\
Number of generations & 8 \\
Max completion length & 512 \\
top-$p$ & 0.99 \\
top-$k$ & 50 \\
Temperature & 0.9 \\
KL coefficient & 0.04 \\
LoRA rank & 8 \\
LoRA alpha & 32 \\
\bottomrule
\end{tabular}
\vspace{-1.5em}
\end{table}

\subsection{Training Details}
    \subsubsection{Implementation details.}
    We report the full hyperparameter settings in Table~\ref{tab:hyperparams}.
    We apply LoRAs~\cite{lora}, with rank 8 and alpha 32, to the LLM of $\pi$ and optimize only the LoRA parameters while keeping all other pre-trained parameters frozen.
    We use AdamW~\cite{adamw} and GRPO~\cite{shao2024deepseekmath} with a group size of $N{=}8$, KL coefficient${=}0.04$, and a learning rate of $\eta{=}1{\times}10^{-6}$, implemented using ms-swift~\cite{msswift}.
    Stages~1--4 are trained for 300, 300, 600, and 900 steps, respectively, and the best-performing checkpoint from each stage is used to initialize the subsequent stage.
    We employ Qwen3-VL-8B as the judge model for the compression fidelity reward $R_\text{Fid}$.
    For the grounding reward, we set $\tau_{\text{iou}}{=}0.5$, $\tau_b{=}10$, and $\tau_p{=}100$.
    All experiments are conducted on a system equipped with 4$\times$ NVIDIA RTX A6000 GPUs with 48\,GB of memory per GPU, an AMD Ryzen Threadripper PRO 3955WX 16-core CPU, and 160\,GB of system memory, running Ubuntu 22.04.5 LTS; the remaining software dependencies and their exact versions are specified in the code.

    \subsubsection{Prompts.}
    \begin{figure*}[!ht]
        \centering
        \begin{subfigure}[b]{\linewidth}
            \includegraphics[width=\linewidth]{Figure/suppl/prompt.pdf}
            \caption{Default prompt}
            \label{fig:training_prompt}
        \end{subfigure}
        \hfill
        \begin{subfigure}[b]{\linewidth}
            \includegraphics[width=\linewidth]{Figure/suppl/prompt_judge.pdf}
            \caption{Judge prompt for Stage~3}
            \label{fig:stage3_judge_prompt}
        \end{subfigure}
        \caption{Prompt for (a) the structured answer block, (b) Compression fidelity in Stage~3.}
        \label{fig:prompts}
    \end{figure*}

    We provide the prompt for the structured answer block in \figref{fig:training_prompt}.
    The judge prompt for compression fidelity (Stage~3) are shown in \figref{fig:stage3_judge_prompt}.

\section{Additional Experiments}

\begin{table*}[!t]
\centering
\small
\caption{Referring expression segmentation on RefCOCO~\cite{refcoco}, RefCOCO+~\cite{refcoco}, and RefCOCOg~\cite{refcocog} benchmarks.
We report the performance (cIoU) evaluated on validation, testA, and testB splits for RefCOCO and RefCOCO+, and validation and test splits for RefCOCOg.
For SAM3, $+$ denotes joint text-and-bounding-box prompting, whereas the standard setting uses text-only prompting.
Excluding the oracle results, the best score in each column is shown in bold.}
\label{tab:refcoco_main_sam3}

\setlength{\tabcolsep}{3pt}
\begin{tabular*}{\textwidth}{
  @{\extracolsep{\fill}}
  lllcccccccc
}
\toprule
\multirow{2}{*}{Method}
& \multirow{2}{*}{MLLM}
& \multirow{2}{*}{Dec.}
& \multicolumn{3}{c}{RefCOCO}
& \multicolumn{3}{c}{RefCOCO+}
& \multicolumn{2}{c}{RefCOCOg} \\
\cmidrule(lr){4-6}
\cmidrule(lr){7-9}
\cmidrule(lr){10-11}
& & &
val & tA & tB
& val & tA & tB
& val & test \\
\midrule

\multicolumn{11}{l}{
\hspace{-6pt}\gray{\textit{Oracle with GT}}
} \\
SAM~\cite{sam}
& -- & SAM
& 82.6 & 84.1 & 81.3
& 82.0 & 84.1 & 81.3
& 80.4 & 81.4 \\

SAM2~\cite{sam2}
& -- & SAM2
& 83.2 & 84.9 & 81.9
& 83.0 & 84.9 & 81.9
& 81.0 & 81.8 \\

SAM3~\cite{sam3}
& -- & SAM3
& 26.4 & 26.5 & 27.4
& 26.4 & 26.5 & 27.5
& 34.3 & 35.1 \\

SAM3$^{+}$~\cite{sam3}
& -- & SAM3$^{+}$
& 76.1 & 75.8 & 75.4
& 76.4 & 75.8 & 75.4
& 72.9 & 73.8 \\
\midrule

\multicolumn{11}{l}{
\hspace{-6pt}\gray{\textit{LLM-7B with SAM}}
} \\
Qwen2.5-Omni-7B~\cite{qwen2.5omni}
& Qw2.5OM-7B & SAM
& 74.6 & 76.6 & 70.1
& 67.1 & 71.5 & 61.3
& 69.8 & 72.0 \\

\methodshort{} (\textbf{Ours})
& Qw2.5OM-7B & SAM
& \textbf{79.9} & \textbf{82.3} & \textbf{74.8}
& \textbf{74.3} & 78.6 & \textbf{66.8}
& 73.2 & \textbf{76.6} \\
\midrule

\multicolumn{11}{l}{
\hspace{-6pt}\gray{\textit{LLM-7B with SAM2}}
} \\
Qwen2.5-Omni-7B~\cite{qwen2.5omni}
& Qw2.5OM-7B & SAM2
& 74.5 & 75.8 & 69.1
& 66.5 & 71.2 & 60.5
& 71.8 & 71.2 \\

\methodshort{} (\textbf{Ours})
& Qw2.5OM-7B & SAM2
& 79.4 & 81.5 & 73.6
& 74.0 & \textbf{78.7} & 66.1
& \textbf{75.3} & 76.0 \\
\midrule

\multicolumn{11}{l}{
\hspace{-6pt}\gray{\textit{LLM-7B with SAM3}}
} \\
Qwen2.5-Omni-7B~\cite{qwen2.5omni}
& Qw2.5OM-7B & SAM3
& 46.1 & 45.2 & 37.5
& 43.5 & 44.8 & 36.4
& 47.9 & 54.2 \\

Qwen2.5-Omni-7B~\cite{qwen2.5omni}
& Qw2.5OM-7B & SAM3$^{+}$
& 73.7 & 68.2 & 64.8
& 64.1 & 65.4 & 55.8
& 67.9 & 71.8 \\

\methodshort{}
& Qw2.5OM-7B & SAM3
& 48.2 & 47.5 & 39.3
& 45.4 & 46.5 & 39.4
& 49.8 & 56.9 \\

\methodshort{}
& Qw2.5OM-7B & SAM3$^{+}$
& 79.3 & 72.2 & 69.1
& 70.1 & 70.1 & 62.1
& 70.7 & 72.7 \\
\midrule

\multicolumn{11}{l}{
\hspace{-6pt}\gray{\textit{LLM-3B with SAM}}
} \\
Qwen2.5-Omni-3B~\cite{qwen2.5omni}
& Qw2.5OM-3B & SAM
& 66.0 & 66.0 & 63.9
& 59.2 & 61.2 & 51.8
& 58.6 & 66.3 \\

\methodshort{}
& Qw2.5OM-3B & SAM
& 74.7 & 79.9 & 67.4
& 66.8 & 71.7 & 59.8
& 67.2 & 76.0 \\
\midrule

\multicolumn{11}{l}{
\hspace{-6pt}\gray{\textit{LLM-3B with SAM2}}
} \\
Qwen2.5-Omni-3B~\cite{qwen2.5omni}
& Qw2.5OM-3B & SAM2
& 65.6 & 65.3 & 62.0
& 58.7 & 61.2 & 50.7
& 60.0 & 65.5 \\

\methodshort{}
& Qw2.5OM-3B & SAM2
& 74.9 & 79.8 & 66.3
& 66.5 & 72.0 & 58.7
& 68.3 & 75.6 \\
\midrule

\multicolumn{11}{l}{
\hspace{-6pt}\gray{\textit{LLM-3B with SAM3}}
} \\
Qwen2.5-Omni-3B~\cite{qwen2.5omni}
& Qw2.5OM-3B & SAM3
& 45.8 & 40.5 & 37.0
& 43.5 & 38.5 & 37.3
& 48.2 & 51.5 \\

Qwen2.5-Omni-3B~\cite{qwen2.5omni}
& Qw2.5OM-3B & SAM3$^{+}$
& 66.7 & 67.0 & 60.1
& 58.9 & 60.5 & 53.3
& 61.5 & 71.4 \\

\methodshort{}
& Qw2.5OM-3B & SAM3
& 48.0 & 42.5 & 38.9
& 45.8 & 40.0 & 39.6
& 49.8 & 53.7 \\

\methodshort{}
& Qw2.5OM-3B & SAM3$^{+}$
& 70.4 & 70.8 & 61.7
& 63.2 & 63.2 & 55.4
& 63.8 & 74.9 \\
\bottomrule
\end{tabular*}
\end{table*}

\subsection{Results on Training Benchmarks}
Extending the comparisons on RES and R-VOS in the main paper, we additionally evaluate TALE~\cite{tale} and ACON~\cite{acon} on ReVOS~\cite{visa} and RefAVS~\cite{refavs}.
As shown in \tabref{tab:revos_refavs}, TCA outperforms TALE by +10.8, +14.6, and +11.1 points and ACON by +9.9, +14.7, and +10.0 points in $\mathcal{J}\&\mathcal{F}$ on ReVOS, RefAVS-S, and RefAVS-U, respectively.
These results further demonstrate the effectiveness of our IB-guided reasoning compression over generic compression methods in both zero-shot and in-domain settings.

    \begin{table*}[t]
\centering
\caption{Referring video object segmentation on the Ref-YouTube-VOS (RY-V)~\cite{refytvos}, Ref-DAVIS17 (RD17)~\cite{refdavis17}, and MeViS~\cite{mevis} benchmarks.
We report the performance ($\mathcal{J}\&\mathcal{F}$) evaluated on the validation set of each benchmark.
$+$ indicates that the pre-trained frozen decoder (Dec.) takes both a text and bounding box coordinates as inputs.
Excluding the oracle results, the best score in each column is shown in bold.}
\label{tab:rvos_main_sam3}
\setlength{\tabcolsep}{6pt}
\small
\begin{tabular}{lll ccc}
\toprule
Method & MLLM & Dec. & RY-V & RD17 & MeViS \\
\midrule
\multicolumn{6}{l}{\hspace{-6pt}\gray{\textit{Oracle with GT}}} \\
SAM2~\cite{sam2} & - & SAM2 & 86.8 & 87.4 & 81.8 \\
SAM3~\cite{sam3} & - & SAM3 & 43.3 & 51.3 & 34.5 \\
SAM3$^{+}$~\cite{sam3} & - & SAM3$^{+}$ & 46.7 & 52.7 & 40.5 \\
\midrule

\multicolumn{6}{l}{\hspace{-6pt}\gray{\textit{LLM-3B with SAM2}}} \\
Qwen2.5-Omni-3B~\cite{qwen2.5omni} & Qw2.5OM-3B & SAM2
& 32.4 & 53.1 & 21.2 \\
\methodshort{} & Qw2.5OM-3B & SAM2
& \textbf{69.9} & \textbf{74.6} & \textbf{52.7} \\
\midrule

\multicolumn{6}{l}{\hspace{-6pt}\gray{\textit{LLM-3B with SAM3}}} \\
Qwen2.5-Omni-3B~\cite{qwen2.5omni} & Qw2.5OM-3B & SAM3
& 22.0 & 31.9 & 16.6 \\
Qwen2.5-Omni-3B~\cite{qwen2.5omni} & Qw2.5OM-3B & SAM3$^{+}$
& 20.9 & 30.2 & 17.3 \\
\methodshort{} & Qw2.5OM-3B & SAM3
& 49.8 & 47.7 & 45.2 \\
\methodshort{} & Qw2.5OM-3B & SAM3$^{+}$
& 47.3 & 45.4 & 44.0 \\
\midrule

\multicolumn{6}{l}{\hspace{-6pt}\gray{\textit{LLM-7B with SAM3}}} \\
Qwen2.5-Omni-7B~\cite{qwen2.5omni} & Qw2.5OM-7B & SAM3
& 48.6 & 39.5 & 39.5 \\
Qwen2.5-Omni-7B~\cite{qwen2.5omni} & Qw2.5OM-7B & SAM3$^{+}$
& 48.6 & 43.6 & 40.4 \\
\methodshort{} & Qw2.5OM-7B & SAM3
& 49.4 & 48.9 & 44.1 \\
\methodshort{} & Qw2.5OM-7B & SAM3$^{+}$
& 50.5 & 48.9 & 44.2 \\
\bottomrule
\end{tabular}
\end{table*}
    \begin{table*}[t]
\centering
\small
\caption{Reasoning video object segmentation on the ReVOS~\cite{visa} benchmark and referring audio-visual segmentation on the RefAVS~\cite{refavs} benchmark.
We report region similarity ($\mathcal{J}$), contour accuracy ($\mathcal{F}$), and their average ($\mathcal{J}\&\mathcal{F}$).
Excluding the oracle results, the best score in each column is shown in bold.}
\label{tab:revos_refavs_sam3}
\setlength{\tabcolsep}{4pt}
\resizebox{\linewidth}{!}{%
\begin{tabular}{lll ccc ccc ccc}
\toprule
\multirow{2}{*}{Method} & \multirow{2}{*}{MLLM} & \multirow{2}{*}{Dec.}
& \multicolumn{3}{c}{ReVOS} & \multicolumn{3}{c}{RefAVS (Seen)} & \multicolumn{3}{c}{RefAVS (Unseen)} \\
\cmidrule(lr){4-6} \cmidrule(lr){7-9} \cmidrule(lr){10-12}
& & & $\mathcal{J}\&\mathcal{F}$ & $\mathcal{J}$ & $\mathcal{F}$
  & $\mathcal{J}\&\mathcal{F}$ & $\mathcal{J}$ & $\mathcal{F}$
  & $\mathcal{J}\&\mathcal{F}$ & $\mathcal{J}$ & $\mathcal{F}$ \\

\midrule
\multicolumn{6}{l}{\hspace{-6pt}\gray{\textit{Oracle with GT}}} \\
SAM~\cite{sam} & - & SAM
& -- & -- & --
& 93.3 & 91.4 & 95.2
& 91.9 & 93.0 & 90.8 \\
SAM2~\cite{sam2} & - & SAM2
& 79.6 & 76.8 & 82.6
& 92.2 & 90.4 & 94.1
& 92.0 & 92.5 & 91.6 \\
SAM3~\cite{sam3} & - & SAM3
& 25.5 & 24.1 & 26.8
& 66.6 & 65.3 & 67.9
& 71.4 & 71.4 & 71.3 \\
SAM3$^{+}$~\cite{sam3} & - & SAM3$^{+}$
& 30.9 & 30.1 & 31.6
& 71.6 & 70.3 & 72.0
& 76.5 & 75.1 & 77.8 \\

\midrule
\multicolumn{6}{l}{\hspace{-6pt}\gray{\textit{LLM-7B with SAM}}} \\
Qwen2.5-Omni-7B~\cite{qwen2.5omni} & Qw2.5OM-7B & SAM
& -- & -- & --
& 34.2 & 31.9 & 36.4
& 63.9 & 64.2 & 63.6 \\
\methodshort{} (\textbf{Ours}) & Qw2.5OM-7B & SAM
& -- & -- & --
& 69.4 & 66.5 & 72.3
& 77.6 & 78.0 & 77.1 \\

\midrule
\multicolumn{6}{l}{\hspace{-6pt}\gray{\textit{LLM-7B with SAM2}}} \\
Qwen2.5-Omni-7B~\cite{qwen2.5omni} & Qw2.5OM-7B & SAM2
& 38.4 & 37.2 & 39.6
& 35.5 & 30.1 & 40.9
& 64.8 & 64.8 & 64.9 \\
\methodshort{} (\textbf{Ours}) & Qw2.5OM-7B & SAM2
& \textbf{60.5} & \textbf{57.7} & \textbf{63.2}
& \textbf{70.7} & \textbf{68.0} & \textbf{73.4}
& \textbf{78.8} & \textbf{78.7} & \textbf{78.8} \\

\midrule
\multicolumn{6}{l}{\hspace{-6pt}\gray{\textit{LLM-7B with SAM3}}} \\
Qwen2.5-Omni-7B~\cite{qwen2.5omni} & Qw2.5OM-7B & SAM3
& 36.2 & 35.7 & 36.6
& 43.7 & 43.4 & 43.9
& 58.6 & 56.8 & 60.3 \\
Qwen2.5-Omni-7B~\cite{qwen2.5omni} & Qw2.5OM-7B & SAM3$^{+}$
& 34.8 & 33.9 & 35.7
& 43.2 & 43.1 & 43.4
& 58.1 & 56.3 & 59.9 \\
\methodshort{} & Qw2.5OM-7B & SAM3
& 42.9 & 40.0 & 45.8
& 60.3 & 58.5 & 62.1
& 64.5 & 64.2 & 64.8 \\
\methodshort{} & Qw2.5OM-7B & SAM3$^{+}$
& 41.8 & 38.9 & 44.7
& 60.0 & 58.1 & 61.9
& 64.4 & 64.2 & 64.6 \\

\midrule
\multicolumn{6}{l}{\hspace{-6pt}\gray{\textit{LLM-3B with SAM}}} \\
Qwen2.5-Omni-3B~\cite{qwen2.5omni} & Qw2.5OM-3B & SAM
& -- & -- & --
& 33.3 & 30.1 & 36.5
& 54.6 & 53.9 & 55.2 \\
\methodshort{} & Qw2.5OM-3B & SAM
& -- & -- & --
& 56.2 & 54.1 & 58.3
& 61.3 & 61.1 & 61.5 \\

\midrule
\multicolumn{6}{l}{\hspace{-6pt}\gray{\textit{LLM-3B with SAM2}}} \\
Qwen2.5-Omni-3B~\cite{qwen2.5omni} & Qw2.5OM-3B & SAM2
& 35.8 & 34.1 & 37.5
& 33.9 & 30.8 & 37.0
& 55.3 & 54.5 & 56.1 \\
\methodshort{} & Qw2.5OM-3B & SAM2
& 52.7 & 49.3 & 56.0
& 58.0 & 55.4 & 60.6
& 62.3 & 61.3 & 63.4 \\

\midrule
\multicolumn{6}{l}{\hspace{-6pt}\gray{\textit{LLM-3B with SAM3}}} \\
Qwen2.5-Omni-3B~\cite{qwen2.5omni} & Qw2.5OM-3B & SAM3
& 20.4 & 18.8 & 22.0
& 40.0 & 38.5 & 41.5
& 41.5 & 41.2 & 41.7 \\
Qwen2.5-Omni-3B~\cite{qwen2.5omni} & Qw2.5OM-3B & SAM3$^{+}$
& 20.3 & 18.8 & 21.9
& 38.2 & 36.7 & 39.7
& 41.4 & 41.2 & 41.5 \\
\methodshort{} & Qw2.5OM-3B & SAM3
& 36.5 & 33.6 & 39.3
& 45.0 & 43.4 & 46.7
& 46.1 & 45.7 & 46.4 \\
\methodshort{} & Qw2.5OM-3B & SAM3$^{+}$
& 36.4 & 33.5 & 39.3
& 43.2 & 41.5 & 44.8
& 46.2 & 45.9 & 46.4 \\

\bottomrule
\end{tabular}%
}
\end{table*}
    
\subsection{Ablation Analysis}

    In this section, we provide additional analyses of TCA, focusing on the LLM and decoder backbones, the optimization stages, and the choice of LLM judge in Stage~3. 

    \subsubsection{Performance with different backbones.}       
    In \tabref{tab:refcoco_main_sam3}, \tabref{tab:revos_refavs_sam3} and \tabref{tab:rvos_main_sam3}, we first introduce the oracle settings for the SAM~\cite{sam}, SAM2~\cite{sam2}, and SAM3~\cite{sam3} decoders.
    Unlike SAM2 and SAM3, SAM does not provide a built-in video tracking mechanism and can only process frames independently.
    We therefore report SAM results only on RefCOCO, which consists of images, and RefAVS, whose videos contain only 10 frames and can be evaluated through frame-wise inference without long-term mask propagation.
    
    To establish the oracle setting, we extract the bounding box coordinates from the ground-truth masks and feed them to the SAM and SAM2 decoders to produce the segmentation masks.
    For SAM3, we use the target label as the text prompt.
    For SAM3$^{+}$, both the text prompt and the bounding box coordinates are provided.
    Since explicit target labels are not available for referring expressions in ReVOS~\cite{visa} and the R-VOS benchmarks~\cite{refytvos,refdavis17,mevis}, we use NLTK~\cite{nltk} to extract the target object noun from the expression and pass it to SAM3 and SAM3$^{+}$.
    Except for the text-only SAM3 settings on ReVOS and R-VOS, whose text prompts are heuristically extracted due to the absence of explicit ground-truth target labels, the remaining oracle results are regarded as the upper bounds attainable with the corresponding decoders.
    The oracle comparison reveals clear performance gaps between SAM/SAM2 and SAM3 across all benchmarks, as shown in the top panel of \tabref{tab:refcoco_main_sam3}, \tabref{tab:rvos_main_sam3} and \tabref{tab:revos_refavs_sam3}.
    We attribute this to the fundamental difference in how each decoder processes prompt conditioning: SAM and SAM2 take bounding box coordinates as primary inputs, which inherently makes mask prediction easier by reducing the discrimination space.
    In contrast, SAM3 fundamentally relies on language descriptions, and even when provided with explicit bounding boxes (SAM3$^{+}$), it does not leverage spatial constraints as effectively as SAM2.
    
    Across different MLLM backbones, TCA-3B with SAM2 consistently outperforms the 3B baseline, improving RY-V, RD17, and MeViS by 37.5, 21.5, and 31.5 points, respectively, and improving ReVOS by 16.9 points in $\mathcal{J}\&\mathcal{F}$.
    Compared with the 7B model under the same SAM2 decoder, TCA-3B drops from 60.5 to 52.7 on ReVOS and from 70.7/78.8 to 58.0/62.3 on RefAVS Seen/Unseen, indicating that reducing the MLLM scale causes a moderate degradation.
    
    Notably, consistent gains are also observed when TCA is paired with SAM.
    These results demonstrate that the improvements of TCAs do not simply arise from replacing SAM with the stronger SAM2 decoder.
    Replacing SAM2 with SAM3, however, causes a substantially larger performance drop than reducing the MLLM scale.
    For the 3B model, TCA decreases from 52.7 to 36.5 on ReVOS and from 58.0/62.3 to 45.0/46.1 on RefAVS Seen/Unseen.
    A similar trend is observed for the 7B model, where the ReVOS performance drops from 60.5 to 42.9.
    SAM3$^{+}$ yields only limited and inconsistent improvements over SAM3 and remains well below SAM2.
    These results show that the effectiveness of TCA is not tied to a specific decoder.
    Instead, the type of prompt consumed by the decoder, particularly box-based versus language-based prompting, has a greater impact on grounding performance than the scale of the MLLM.

    \subsubsection{Performance with different LLM judges.}
    In the main experiments, we employ Qwen3-VL-8B~\cite{qwen3} as the LLM judge model in Stage~3.
    We additionally train TCA with Qwen2.5-VL-7B, Qwen2.5-VL-32B~\cite{qwen2.5vl}, and Llama-3.2-Vision-11B~\cite{llamavi3.2}, and evaluate them on the R-VOS, ReVOS, and RefAVS benchmarks, as shown in \tabref{tab:s2_ablation}.
    Qwen3-VL-8B achieves the best overall performance across the evaluated benchmarks, while Qwen2.5-VL-7B and Qwen2.5-VL-32B yield comparable results.
    Llama-3.2-Vision-11B shows relatively lower performance, particularly on Ref-DAVIS17, ReVOS, and RefAVS.
    These results indicate that TCA remains effective across different judge models, while increasing the judge scale does not consistently improve grounding performance.

    \subsubsection{Performance with different compression types.}
    
    We further investigate the importance of the information retained in the compressed context by comparing reasoning compression with vague and random words.
    As shown in \tabref{tab:ablation_compression}, reasoning compression consistently achieves the best performance across all three benchmarks.
    Replacing the compressed context with vague words decreases $\mathcal{J}\&\mathcal{F}$ by 3.8, 4.4, and 4.5 points on Ref-YouTube-VOS, Ref-DAVIS17, and MeViS, respectively, while random words lead to larger performance drops of 8.6, 9.8, and 9.6 points.
    These results suggest that the benefits of reasoning compression arise not merely from introducing an additional intermediate context, but from preserving informative, target-relevant cues that facilitate accurate grounding.    

    \begin{table}[t]
\centering
\caption{Ablation on the LLM judge.
We report $\mathcal{J}\&\mathcal{F}$ on the referring video object segmentation benchmarks, the RefAVS Unseen test split, and the ReVOS validation split.
We use SAM2 as the decoder.
The best score in each column is shown in bold.}
\label{tab:s2_ablation}
\small
\setlength{\tabcolsep}{4pt}
\begin{tabular}{lccccc}
\toprule
LLM Judge & RY-V & RD17 & MeViS & ReVOS & RefAVS(U) \\
\midrule
Qw3VL-8B
& \textbf{72.3} & \textbf{78.9} & \textbf{55.8} & \textbf{60.5} & \textbf{78.8} \\
\midrule
Qw2.5VL-7B
& 71.9 & 78.1 & 54.5 & 58.5 & 77.6 \\
Qw2.5VL-32B
& 71.4 & 78.5 & 53.6 & 58.7 & 77.2 \\
Llama3.2Vi-11B
& 70.5 & 74.6 & 52.9 & 57.6 & 76.8 \\
\bottomrule
\end{tabular}
\end{table}
    \begin{table}[!t]
\centering
\caption{
    Ablation on compression content.
    We compare reasoning compression with vague and random words as compressed contexts.
    We report $\mathcal{J}\&\mathcal{F}$ on the R-VOS benchmarks.
}
\label{tab:ablation_compression}
\small
\setlength{\tabcolsep}{4pt}
\begin{tabular}{lccccc}
\toprule
Compression & RY-V & RD17 & MeViS  \\
\midrule
Reasoning compression
& \textbf{72.3} & \textbf{78.9} & \textbf{55.8} \\
\midrule
Vague word
& 68.5 & 74.5 & 51.3 \\
Random word
& 63.7 & 69.1 & 46.2 \\
\bottomrule
\end{tabular}
\end{table}

    \subsubsection{Optimization stage.}
    
    \begin{figure*}[t]
        \centering
        \begin{subfigure}[t]{0.32\textwidth}
            \centering
            \includegraphics[width=\linewidth]{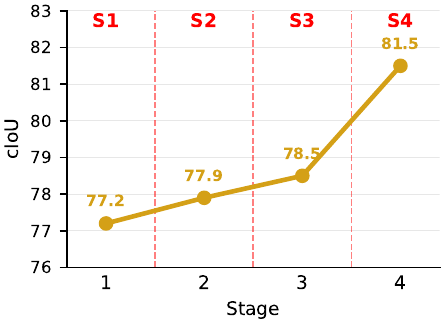}
            \caption{RefCOCO (testA)}
            \label{fig:stage_refcoco}
        \end{subfigure}
        \hfill
        \begin{subfigure}[t]{0.32\textwidth}
            \centering
            \includegraphics[width=\linewidth]{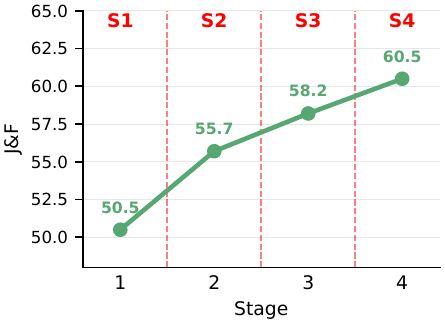}
            \caption{ReVOS}
            \label{fig:stage_rvos}
        \end{subfigure}
        \hfill
        \begin{subfigure}[t]{0.32\textwidth}
            \centering
            \includegraphics[width=\linewidth]{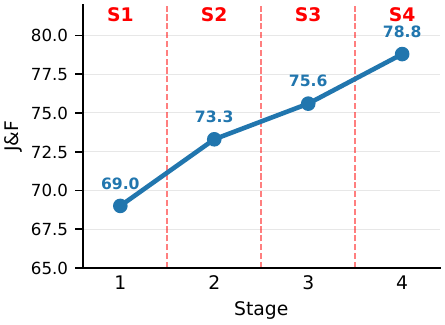}
            \caption{RefAVS (Unseen)}
            \label{fig:stage_refavs}
        \end{subfigure}
            \caption{Performance at each stage on (a) RefCOCO testA, (b) ReVOS, and (c) RefAVS Unseen.}
        \label{fig:stage_all}
    \end{figure*}

    \begin{figure*}[t]
        \centering
    
        \begin{subfigure}[t]{0.47\textwidth}
            \centering
            \includegraphics[width=\linewidth]{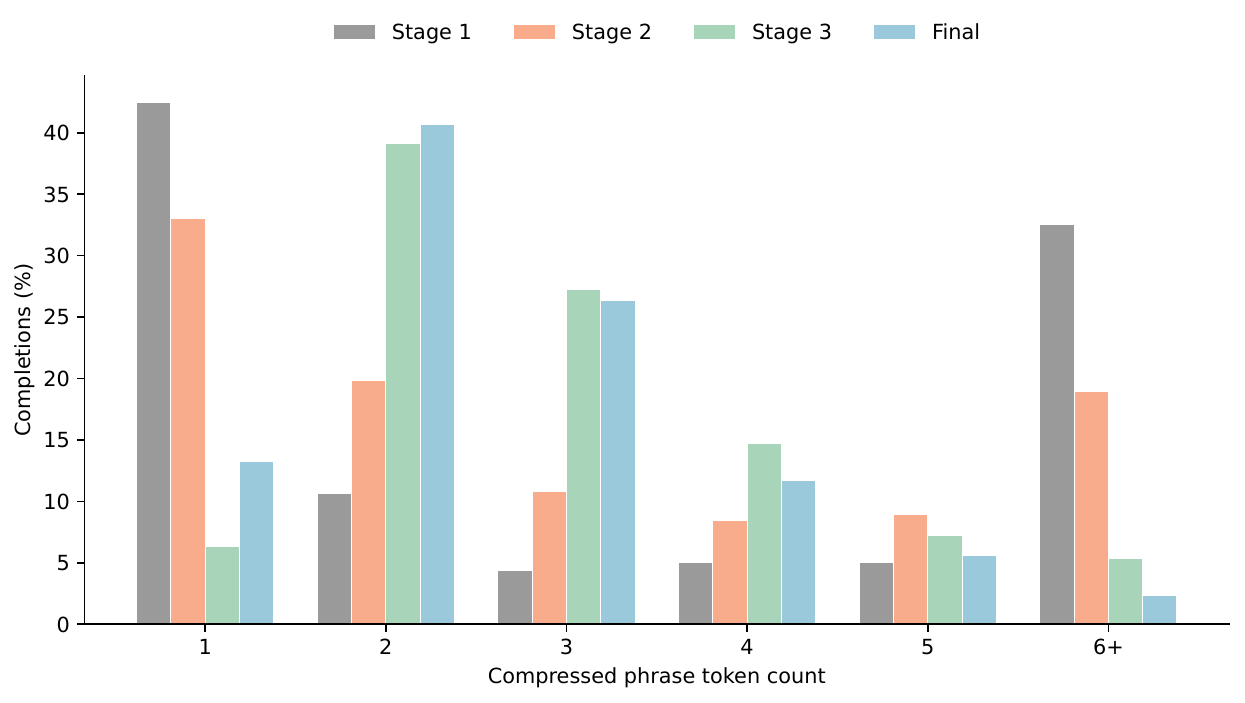}
            \caption{Compressed context token distribution.}
            \label{fig:compression_histogram}
        \end{subfigure}
        \hfill
        \begin{subfigure}[t]{0.25\textwidth}
            \centering
            \includegraphics[width=\linewidth]{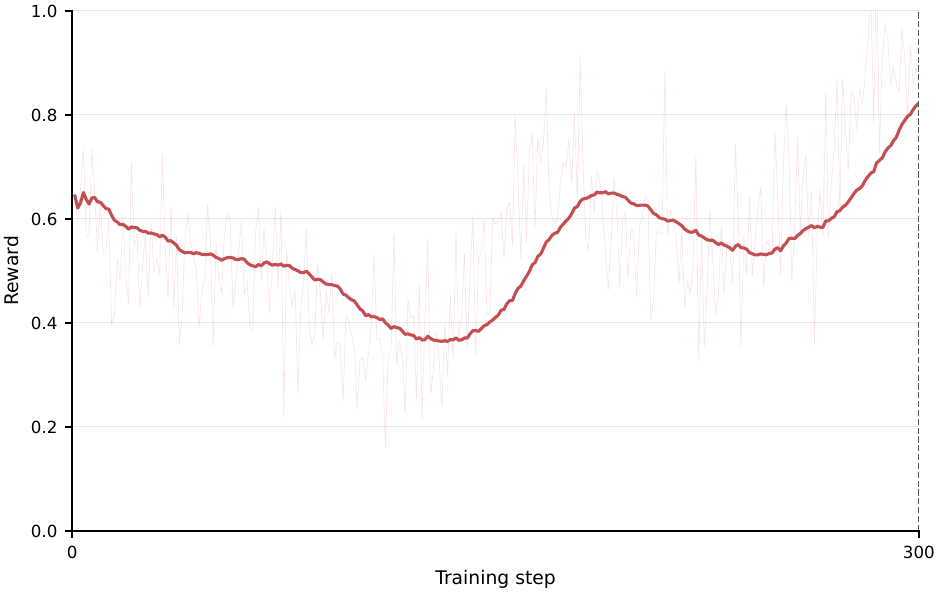}
            \caption{Stage~2 reward.}
            \label{fig:stage2_reward}
        \end{subfigure}
        \hfill
        \begin{subfigure}[t]{0.25\textwidth}
            \centering
            \includegraphics[width=\linewidth]{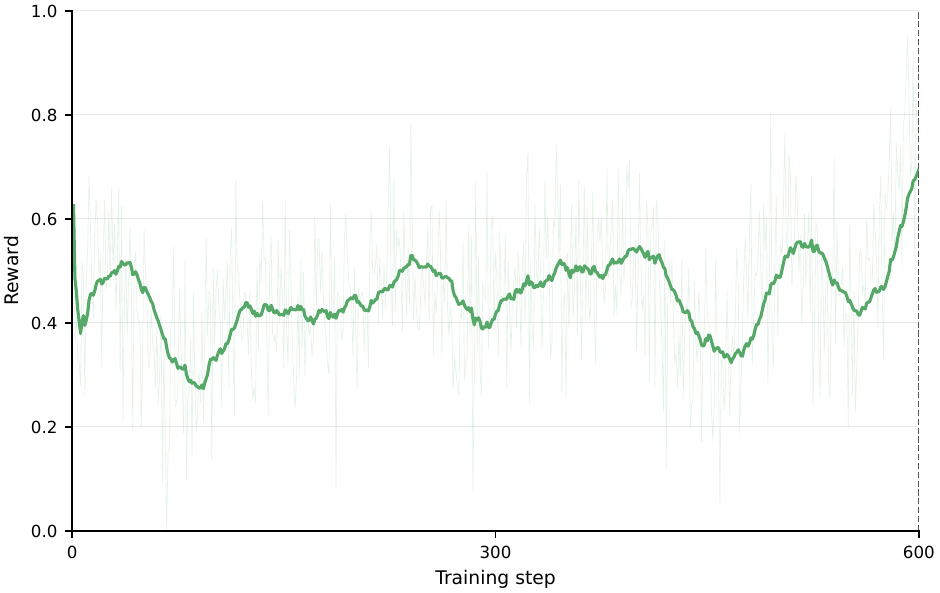}
            \caption{Stage~3 reward.}
            \label{fig:stage3_reward}
        \end{subfigure}
    
        \caption{Compressed-context token distribution and reward dynamics. 
        (a) Distribution of compressed-context lengths across optimization stages.
        (b) Compression-rate reward during Stage~2.
        (c) Compression-fidelity reward during Stage~3.}
        \label{fig:compression_length}
    \end{figure*}
    
    \begin{table}[t]
\centering
\caption{Performance with respect to variants of the stages on the referring expression segmentation benchmarks. We report cIoU evaluated on validation, testA, and testB splits for RefCOCO and RefCOCO+, and validation and test splits for RefCOCOg. We use SAM2 as the decoder.}
\label{tab:refcoco_ablation}
\small
\setlength{\tabcolsep}{2.8pt}
\renewcommand{\arraystretch}{0.8}
\begin{tabular}{cccccccccccc}
\toprule
\multirow{2}{*}{S1} & \multirow{2}{*}{S2} & \multirow{2}{*}{S3} & \multirow{2}{*}{S4}
& \multicolumn{3}{c}{RefCOCO} & \multicolumn{3}{c}{RefCOCO+} & \multicolumn{2}{c}{RefCOCOg} \\
\cmidrule(lr){5-7}\cmidrule(lr){8-10}\cmidrule(lr){11-12}
& & & & val & tA & tB & val & tA & tB & val & test \\
\midrule
\xmark & \xmark & \xmark & \xmark & 74.5 & 75.8 & 69.1 & 66.5 & 71.2 & 60.5 & 71.8 & 71.2 \\
\cmark & \xmark & \xmark & \xmark & 75.6 & 77.2 & 70.6 & 68.2 & 73.5 & 61.8 & 72.9 & 72.1 \\
\cmark & \cmark & \xmark & \xmark & 76.2 & 77.9 & 71.0 & 68.4 & 74.1 & 62.4 & 73.2 & 72.8 \\
\cmark & \cmark & \cmark & \xmark & 76.8 & 78.5 & 70.8 & 69.5 & 74.8 & 63.1 & 72.9 & 73.3 \\
\cmark & \xmark & \xmark & \cmark & 77.5 & 79.6 & 71.6 & 70.8 & 76.2 & 64.2 & 73.9 & 73.8 \\
\cmark & \cmark & \xmark & \cmark & 78.4 & 80.5 & 72.5 & 71.9 & 77.4 & 65.1 & 74.6 & 74.4 \\
\cmark & \cmark & \cmark & \cmark & \textbf{79.4} & \textbf{81.5} & \textbf{73.6} & \textbf{74.0} & \textbf{78.7} & \textbf{66.1} & \textbf{75.3} & \textbf{76.0} \\
\bottomrule
\end{tabular}
\end{table}

\begin{table}[t]
\centering
\caption{Performance with respect to variants of the stages on ReVOS and RefAVS. We report $\mathcal{J}\&\mathcal{F}$ on the ReVOS validation set and the RefAVS Unseen test split. We use SAM2 as the decoder.}
\label{tab:revos_refavs_ablation}
\small
\setlength{\tabcolsep}{6pt}
\renewcommand{\arraystretch}{0.8}
\begin{tabular}{@{}cccccc@{}}
\toprule
\multirow{2}{*}{S1}
& \multirow{2}{*}{S2}
& \multirow{2}{*}{S3}
& \multirow{2}{*}{S4}
& \multicolumn{2}{c}{$\mathcal{J}\&\mathcal{F}$} \\
\cmidrule(lr){5-6}
& & & & ReVOS & RefAVS (U) \\
\midrule
\xmark & \xmark & \xmark & \xmark & 38.4 & 64.8 \\
\cmark & \xmark & \xmark & \xmark & 50.5 & 69.0 \\
\cmark & \cmark & \xmark & \xmark & 55.7 & 73.3 \\
\cmark & \cmark & \cmark & \xmark & 58.2 & 75.6 \\
\cmark & \xmark & \xmark & \cmark & 56.6 & 74.9 \\
\cmark & \cmark & \xmark & \cmark & 58.1 & 76.4 \\
\cmark & \cmark & \cmark & \cmark & \textbf{60.5} & \textbf{78.8} \\
\bottomrule
\end{tabular}
\end{table}
    
    We train the model by ablating each stage and evaluate the performance on the RES, ReVOS, and RefAVS benchmarks as shown in  \tabref{tab:refcoco_ablation},  \tabref{tab:revos_refavs_ablation}.
    As discussed in the main paper, the compression step (\ie, Stage~2) is consistently effective and brings substantial improvements across all benchmarks.
    In \figref{fig:stage_all}, we depict the grounding performance evaluated after each stage on RefCOCO, ReVOS, and RefAVS.
    The results on RefCOCO show that each stage leads to performance improvements despite the zero-shot setting.
    At the same time, reasoning-context compression (Stage~2) on ReVOS and RefAVS introduces the steepest performance gain, highlighting the effectiveness of reasoning compression in multimodal grounding.
    
    \subsubsection{Compression length.}

    The primary goal of TCA is to compress extended multimodal reasoning into a concise, target-specific context that preserves only the information necessary for multimodal grounding.
    To analyze the distribution of compressed-context lengths throughout optimization, we depict the token distribution at each optimization stage.

    As shown in \figref{fig:compression_histogram}, the compressed context produced by the model trained only with the format reward are concentrated at two extremes: overly short single-token contexts and verbose outputs containing six or more tokens.
    In Stage~2, optimization with the compression-rate reward progressively shortens verbose contexts while expanding single-token contexts into more informative representations.
    In Stage~3, the compression-fidelity reward further refines these contexts by encouraging them to retain sufficient target-relevant evidence for accurate prediction, thereby reducing both under-compressed and over-compressed contexts.
    The grounding reward in Stage~4 then optimizes the retained information according to the final grounding objective instead of substantially altering the context length.
    In practice, the performance comparison in \figref{fig:stage_all} and the token length distribution in \figref{fig:compression_histogram} show that the overall compressed-context length distribution is largely established by Stage~3, whereas Stage~4 completes the task-specific refinement of the compressed contexts.
        
    As shown in \figref{fig:stage2_reward} and \figref{fig:stage3_reward}, both the compression-rate and compression fidelity rewards show an overall increasing trend during training.
    This reward improvement is consistent with the distributional changes observed in \figref{fig:compression_histogram}, indicating that the model learns to balance context compactness and target-relevant fidelity rather than merely shortening the output.
    These results verify that the objectives of Stages~2 and 3 are effectively optimized during training.

\subsubsection{Qualitative results.}
\begin{figure*}[p]
    \centering

    \begin{subfigure}[t]{0.48\textwidth}
        \centering
        \includegraphics[width=\linewidth]{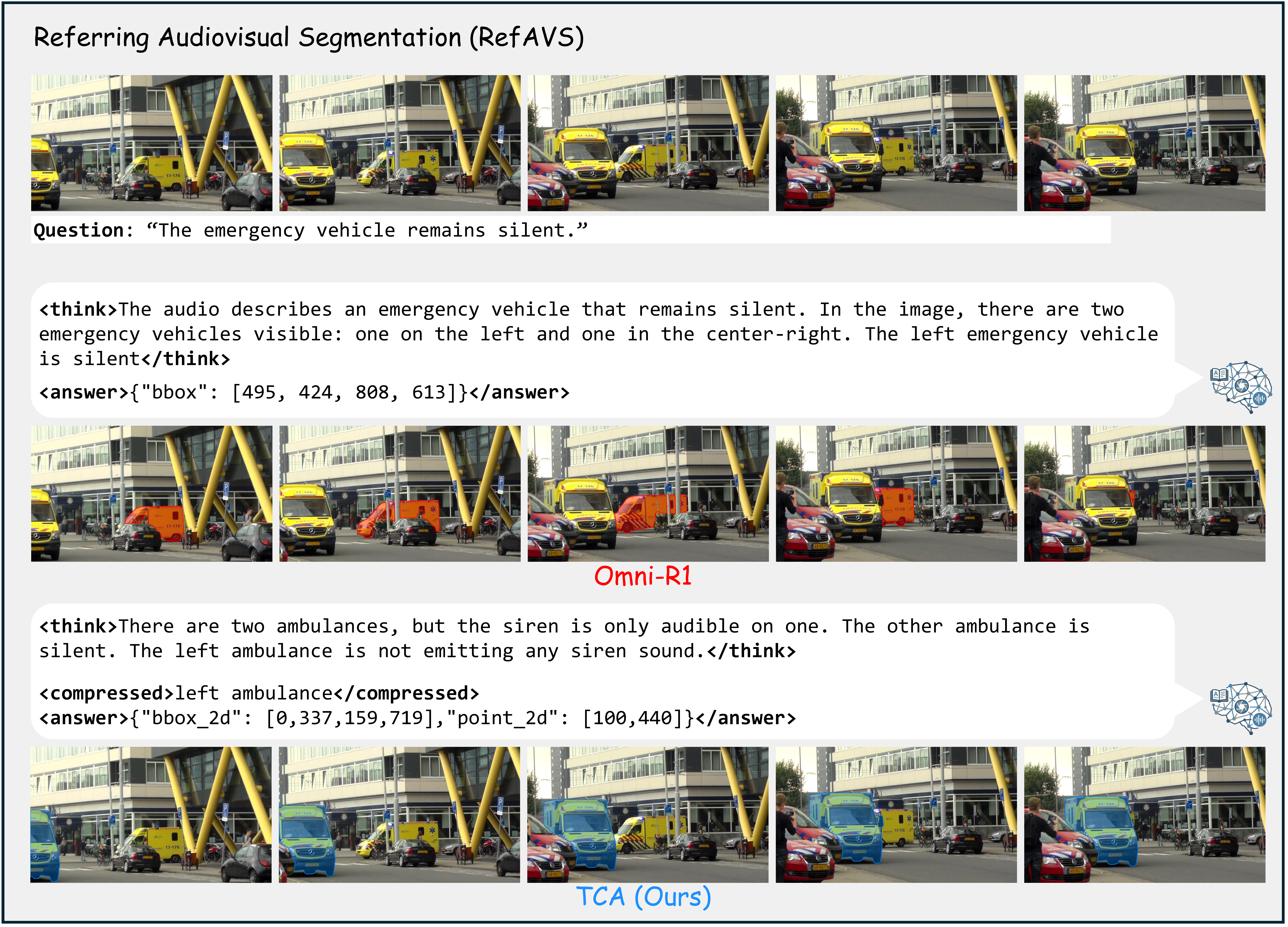}
        \caption{Referring audio-visual segmentation example from RefAVS}
        \label{fig:qualitative_refavs}
    \end{subfigure}
    \hfill
    \begin{subfigure}[t]{0.48\textwidth}
        \centering
        \includegraphics[width=\linewidth]{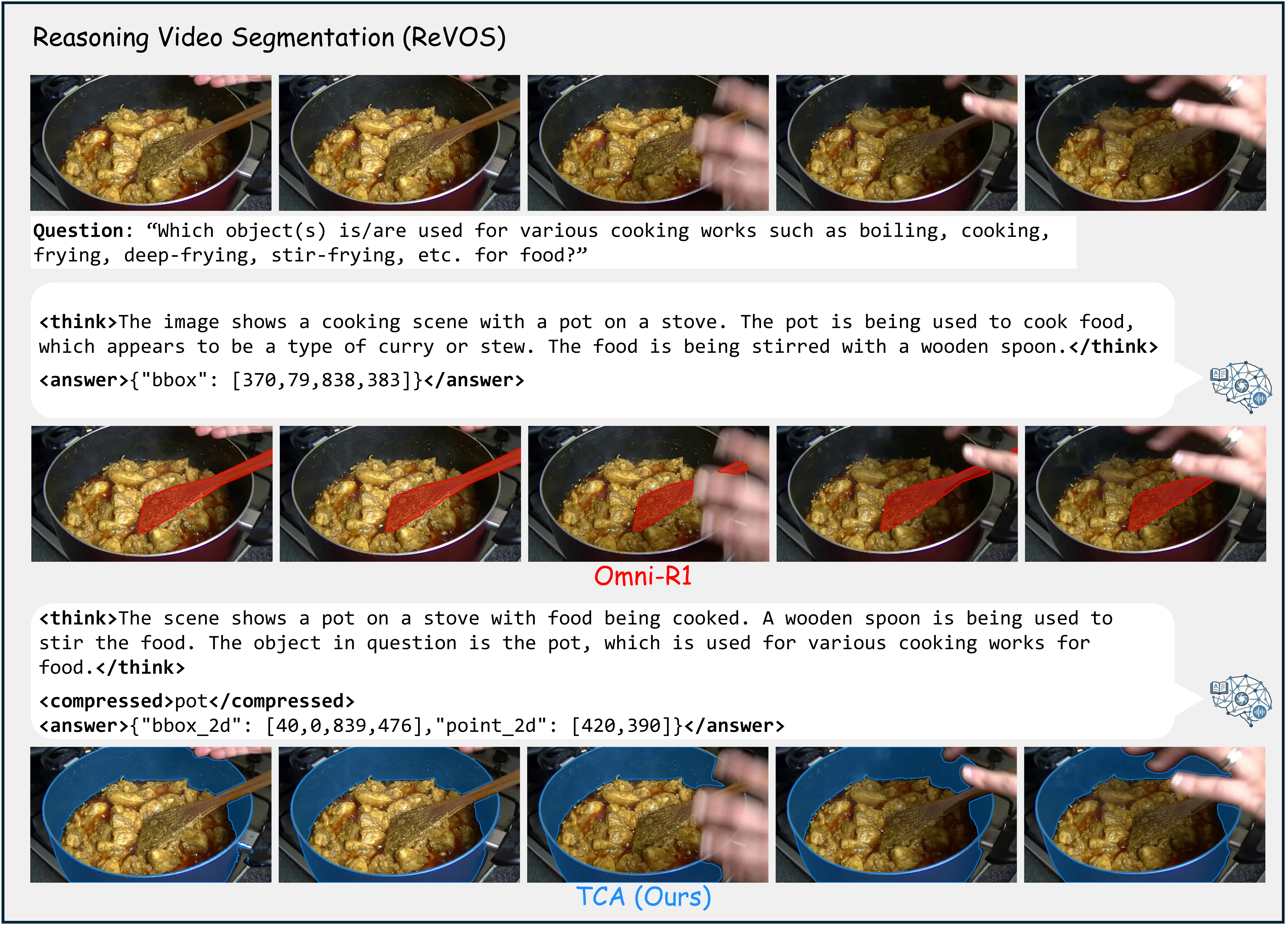}
        \caption{Referring video object segmentation example from ReVOS}
        \label{fig:qualitative_revos}
    \end{subfigure}

    \vspace{0.5em}

    \begin{subfigure}[t]{0.48\textwidth}
        \centering
        \includegraphics[width=\linewidth]{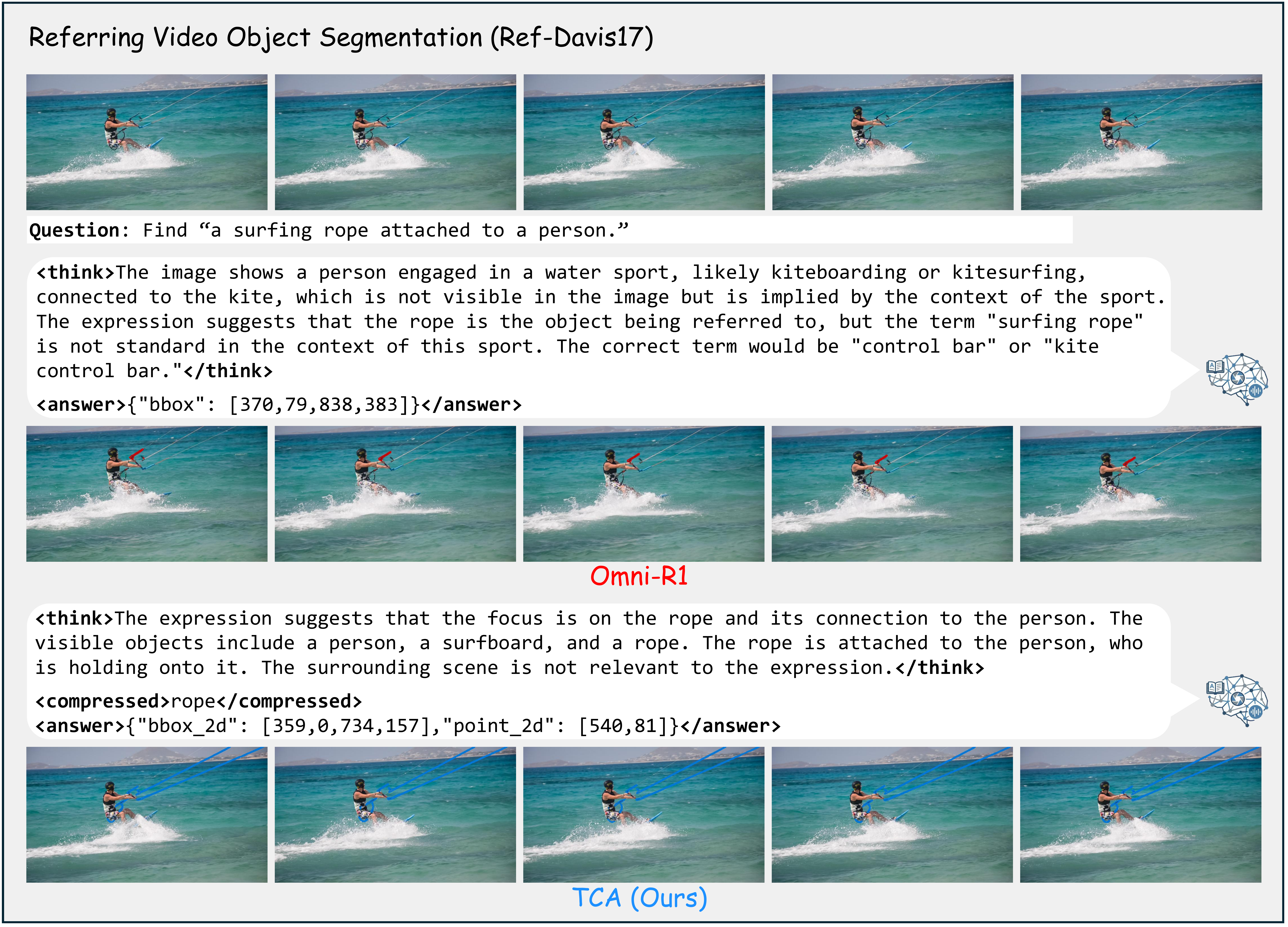}
        \caption{Referring video object segmentation example from Ref-DAVIS17}
        \label{fig:qualitative_refdavis}
    \end{subfigure}
    \hfill
    \begin{subfigure}[t]{0.48\textwidth}
        \centering
        \includegraphics[width=\linewidth]{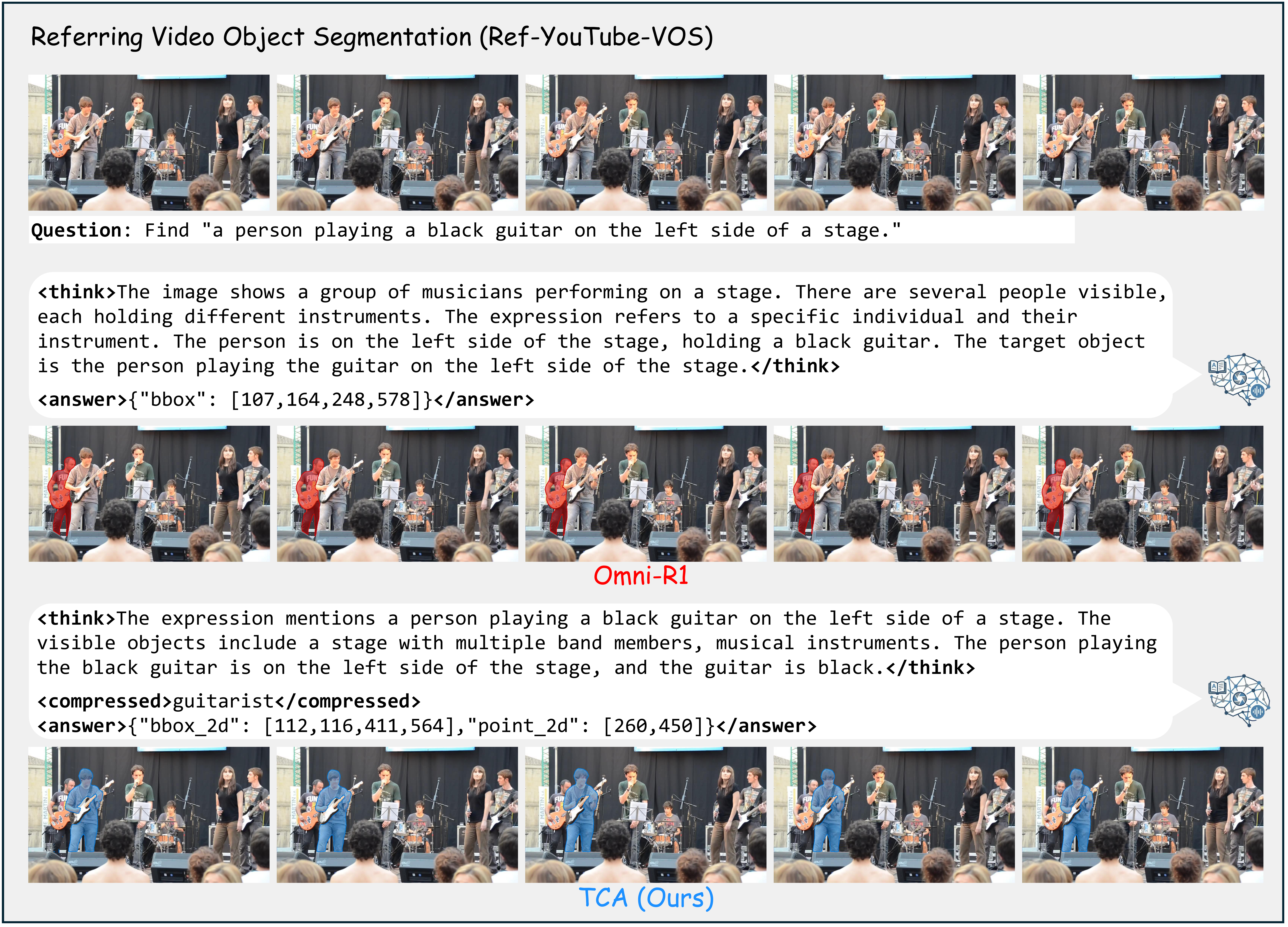}
        \caption{Referring video object segmentation example from Ref-YouYube-VOS}
        \label{fig:qualitative_ytvos}
    \end{subfigure}

    \caption{Qualitative examples on RefAVS, ReVOS and R-VOS benchmarks.}
    \label{fig:qualitative}
\end{figure*}

    In \figref{fig:qualitative}, we present additional qualitative comparisons between TCA and Omni-R1~\cite{omnir1} on RefAVS, ReVOS, and R-VOS benchmarks.
    The example in \figref{fig:qualitative_refavs} highlights the importance of reasoning compression in multimodal grounding.
    Omni-R1 produces an incorrect bounding box despite identifying the correct cue (\eg, ``left ambulance''), whereas TCA transfers this cue into a concise compressed context that remains semantically consistent with the reasoning and the grounding objective, resulting in accurate localization.
    The example in \figref{fig:qualitative_revos} shows that, while both methods reason about the correct target, Omni-R1 yields a less precise bounding box, whereas TCA compresses the reasoning into a target-aware context (\eg, ``pot''), resulting in tighter localization and more accurate segmentation.
    In \figref{fig:qualitative_refdavis}, Omni-R1 drifts from the referred object (\ie, ``surfing rope'') to a related object (\ie, ``handle''), leading to incorrect localization, whereas TCA preserves the target cue in a concise compressed context and accurately localizes the surfing rope.
    In \figref{fig:qualitative_ytvos}, Omni-R1 localizes a neighboring musician despite reasoning about the correct target, whereas TCA compresses the reasoning into a concise target-specific context (\ie, ``guitarist''), resulting in accurate localization and temporally consistent segmentation.

    \subsubsection{Impact of reasoning compression.}
\begin{figure*}[p]
        \centering
        \includegraphics[width=\linewidth]{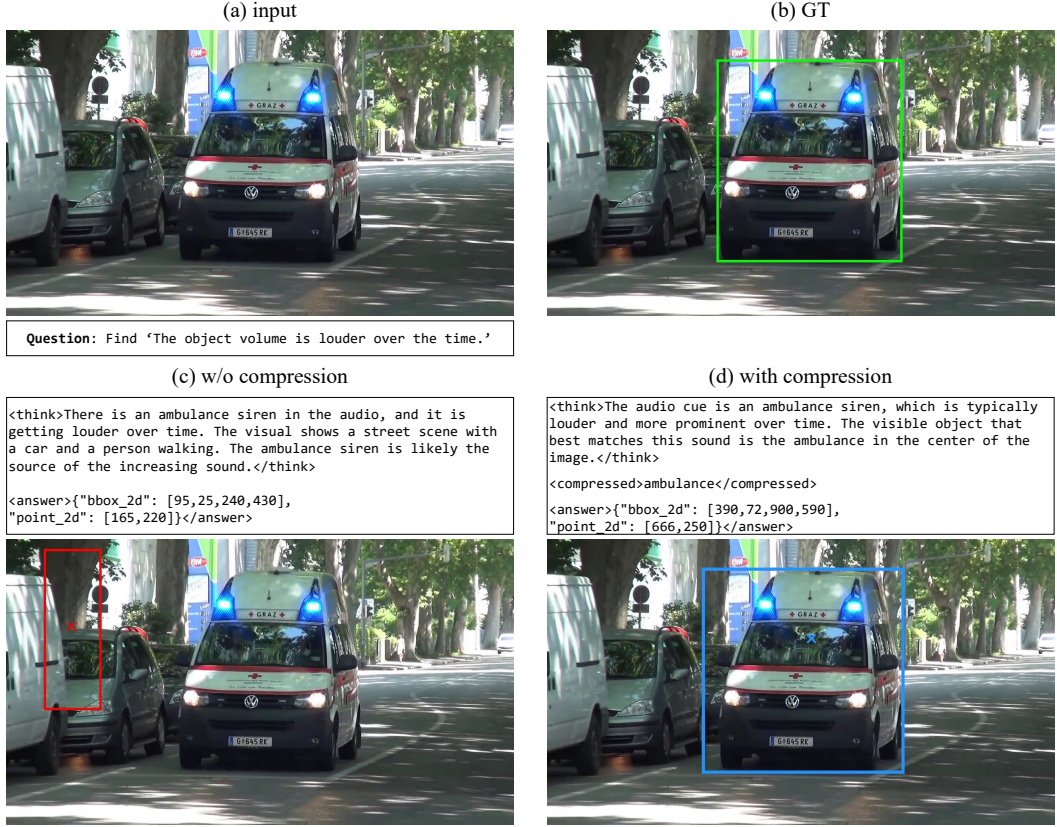}    
        \caption{Impact of compression. TCA extracts target-relevant context from the reasoning path, yielding a more precise mask.}
        \label{fig:compression_vs_non}
    \end{figure*}

    We present another example to further validate the impact of reasoning compression in \figref{fig:compression_vs_non}.
    Without compression, the model often fails to identify the key context from long reasoning paths, leading to an inaccurate localization prediction.
    In our TCA, $\mathbf{c}$ retains only target-relevant cues, producing a more precise segmentation.

    \subsubsection{Failure case.}
\begin{figure*}[p]
        \includegraphics[width=\linewidth]{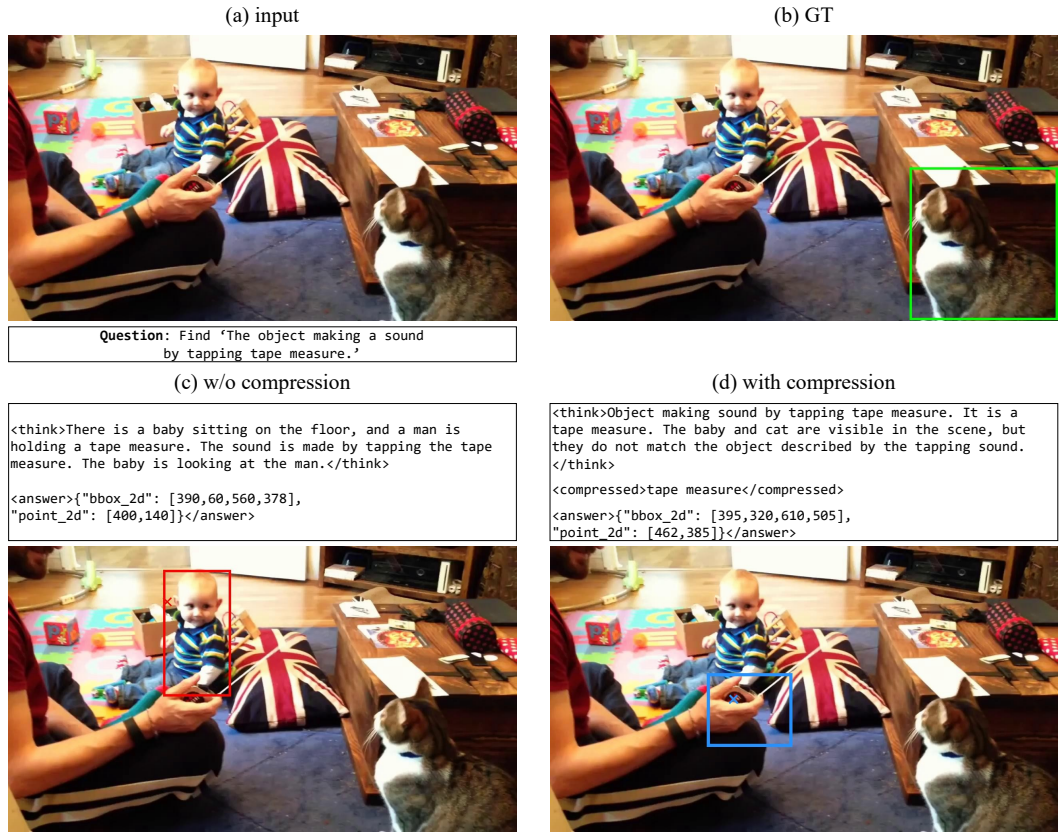}    
        \caption{Failure case from RefAVS. When the reasoning phase identifies a related but incorrect object, compression preserves the erroneous target context and cannot recover the correct localization.}
        \label{fig:failure_case}
    \end{figure*}

    Our approach assumes that the target-relevant context for multimodal grounding can be extracted from extended reasoning paths. 
    This assumption has an inherent limitation: when the reasoning phase itself is incorrect, compression may faithfully preserve the wrong context and thus cannot recover the correct target. 
    We illustrate a failure case from RefAVS in \figref{fig:failure_case}. 
    In the example, the model compresses reasoning as intended, but the reasoning phase misidentifies the target object as the tape measure, although the referred object is the cat interacting with it.
    While several results suggest that TCA implicitly improves the reasoning behavior of MLLMs, as shown in Fig.~5 of the main paper and \figref{fig:qualitative}, this benefit is not guaranteed when the underlying reasoning is incorrect.
    This limitation highlights that TCA primarily mitigates drift by compressing intermediate content, rather than correcting fundamentally wrong reasoning, and motivates future work on stronger verification or correction mechanisms within the reasoning phase.
    
\end{document}